\documentclass{article}

\usepackage[final]{neurips_2025}

\usepackage[utf8]{inputenc} 
\usepackage[T1]{fontenc}    
\usepackage[colorlinks=true,
linkcolor=blue,
citecolor=blue,
urlcolor=blue]{hyperref}     
\usepackage{url}            
\usepackage{booktabs}       
\usepackage{amsfonts}       
\usepackage{nicefrac}       
\usepackage{comment}
\usepackage{microtype}      
\usepackage{xcolor}         
\usepackage{graphicx}
\usepackage{amsmath}
\usepackage{amssymb}
\usepackage{mathtools}
\usepackage{amsthm}
\usepackage{mathrsfs}
\usepackage{bm}
\usepackage{nccmath}   
\usepackage{float}
\usepackage{algorithm}
\usepackage{algorithmic}
\usepackage{caption}
\usepackage{subfigure}
\usepackage{subcaption} 

\newcommand{\R}{\mathbb{R}}

\newcommand{\E}{\mathbb{E}}

\DeclareMathOperator*{\argmax}{arg\,max}
\DeclareMathOperator*{\argmin}{arg\,min}

\newcommand{\crl}[1]{\ensuremath{ \left\{ #1 \right\} }}
\newcommand{\edg}[1]{\ensuremath{\! \left[ #1 \right] }}
\newcommand{\brak}[1]{\ensuremath{\left( #1 \right)}}

\theoremstyle{plain}
\newtheorem{theorem}{Theorem}[section]
\newtheorem{proposition}[theorem]{Proposition}

\theoremstyle{definition}

\theoremstyle{remark}

\title{Deep Learning for  Continuous-Time\\ Stochastic  Control with Jumps}

\author{%
 Patrick Cheridito\\
  Department of Mathematics\\
  ETH Zurich, Switzerland 
   \And
   Jean-Loup Dupret\thanks{Corresponding author.
    }  \\
   Department of Mathematics\\
   ETH Zurich, Switzerland 
   \And
  Donatien Hainaut \\
  LIDAM-ISBA \\
    UCLouvain, Belgium 
    \AND \\[-0.7cm]
\texttt{ \{patrickc,jdupret\}@ethz.ch, \{donatien.hainaut\}@uclouvain.be}
}

\begin{document}

\maketitle
	\hypersetup{
	linkcolor=blue,
	urlcolor=blue,
	citecolor=black,
}
\begin{abstract}
In this paper, we introduce a model-based deep-learning approach to solve finite-horizon continuous-time 
 stochastic control problems with jumps. We iteratively  train two neural networks: one to represent the optimal policy and the other to approximate the value function. Leveraging a continuous-time version of the dynamic programming principle, we derive two different training objectives based on the Hamilton--Jacobi--Bellman equation, ensuring that the networks capture the underlying stochastic dynamics. Empirical evaluations on different problems illustrate the accuracy and scalability of our approach, demonstrating its effectiveness in solving complex high-dimensional stochastic 
control tasks. Code is available at \url{https://github.com/jdupret97/Deep-Learning-for-CT-Stochastic-Control-with-Jumps}. 
\end{abstract}

\section{Introduction} 
\label{sec:intro}
A  large class of dynamic decision-making problems under uncertainty can be  modeled as continuous-time  stochastic control problems. In this paper, we introduce two neural network-based numerical algorithms for such problems in high dimensions with finite time horizon and jumps. More precisely, we consider control problems of the form \vspace{-1mm}
 \begin{equation} \label{pr} 
\sup_{\alpha} \E \edg{\int_0^T f(t, X^\alpha_t, \alpha_t) dt + F(X^\alpha_T)} , 
\end{equation}
for a finite horizon $T > 0$, where the supremum is over predictable control 
processes $\alpha = (\alpha_t)_{0 \le t \le T}$ taking\hspace{-0.3mm}  values\hspace{-0.3mm}  in\hspace{-0.3mm}  a\hspace{-0.3mm}  subset\hspace{-0.3mm}  $A \subseteq \R^m$.\ 
The\hspace{-0.3mm}  controlled\hspace{-0.3mm}  process\hspace{-0.3mm}  $X^\alpha$\hspace{-0.3mm}  evolves\hspace{-0.3mm}  in a subset $D \subseteq \R^d$\hspace{-0.25mm} according\hspace{-0.2mm}  to
\begin{equation} \label{sde}
\hspace{-1mm}	dX^\alpha_t = \beta(t, X^\alpha_t, \alpha_t) dt + \sigma(t, X^\alpha_t, \alpha_t) dW_t  + \int_E \hspace{-1mm} \gamma(t, X^\alpha_{t-}, z, \alpha_t) N^{\alpha}(dz, dt), \ \  X^\alpha_0 = x \in D, \hspace{-1mm}
\end{equation}
for an initial condition $x \in D$, an $n$-dimensional Brownian motion $W$ and
a controlled random measure $N^{\alpha}$ on $E \times \R_+$, with $E = \R^l \setminus \{0\}$,
and suitable functions $\beta \colon [0,T] \times D \times A \to \R^d$, 
$\sigma \colon [0,T] \times D \times A \to \R^{d \times n}$ and
$\gamma \colon [0,T] \times D \times E \times A \to \R^d$. 
The functions $f \colon [0,T] \times D \times A \to \R$ 
and $F \colon D \to \R$ model the running and final rewards, respectively. We 
assume the controlled random measure is given by $N^{\alpha}(B \times [0,t]) = \sum_{j =1}^{M^\alpha_t} 1_{\crl{Z_j \in B}}$ for measurable subsets $ B \subseteq E$, where $M^\alpha$ is a Cox process with a 
stochastic intensity of the form
$\lambda(t, X^\alpha_{t-}, \alpha_t)$ and $Z_1, Z_2, \dots$ are i.i.d.\ $E$-valued random vectors
such that, conditionally on $\alpha$, the random elements $W, M^\alpha$ and $Z_1, Z_2, \dots$ are independent.  
Our goal is to find an optimal  control $\alpha^*$ and the corresponding value of problem 
\eqref{pr}. In view of the Markovian nature of the dynamics \eqref{sde}, we work with feedback controls 
of the form $\alpha_t = \alpha(t, X^\alpha_{t-})$ for measurable functions $\alpha \colon [0,T) \times D \to A$
and consider the value function $V: [0,T] \times D \to \mathbb{R}$ given by
\begin{equation*} \label{vf} 
V(t,x)  =  \sup_{\alpha} \E \edg{\int_t^T f(s, X^\alpha_s, \alpha_s) ds + F(X^\alpha_T) \, \Big| \, X^\alpha_t = x} .
\end{equation*} Under suitable assumptions\footnote{see e.g. \cite{soner1988optimal}}, $V$ is the unique solution of the following Hamilton--Jacobi--Bellman (HJB) equation 
\begin{equation}
\partial_t V(t, x) + \sup_{a \in A} H(t, x, V, a)=0, \quad V(T,x) = F(x) \, ,
\label{HJB} 
\end{equation}
for the Hamiltonian $H : [0,T) \times D \times \mathcal{V}  \times A \rightarrow \mathbb{R}$ given\footnote{By $\mathcal{V}$ we denote the set of all functions in $C^{1,2}([0,T) \times \R^d)$
such that the expectation in \eqref{Ham} is finite for all $t,x$ and $a$.}  by   
\begin{equation}  \label{Ham}
\begin{aligned}
H(t,x,V, a) &:=   f(t,x,a)   + \beta^T(t,x,a) \nabla_x V(t,x) + \frac{1}{2} {\rm Tr} 
\edg{\sigma \sigma^T(t,x,a) \nabla^2_x V(t,x)}
\\ &  \hspace{22mm}+ \lambda(t,x,a) \, \E \edg{V (t, x + \gamma(t,x,Z_1,a)) - V(t,x)} \, ,
\end{aligned}
\end{equation} where $\nabla_xV$, $\nabla^2_xV$ denote the gradient and Hessian of $V$  with respect to $x$.  Our approach consists in iteratively training two neural networks to approximate 
the value function $V$ and an optimal control  $\alpha^*$ attaining $V$. It has the following features\vspace{-1mm}:
\begin{itemize}
\item It yields accurate results for high-dimensional continuous-time stochastic control
problems in cases where the underlying system dynamics are known.  \vspace{-0.5mm}
\item It can effectively handle a combination of diffusive noise and random jumps with controlled  intensities. \vspace{-0.5mm}
\item	It can handle general situations where the optimal control is not available in closed form but has to be learned together with the value function.  \vspace{-0.5mm}
\item It approximates both the value function and optimal control at any  point $(t,x) \in [0,T) \times D$\vspace{-1mm}. 
\end{itemize}
While methods like finite differences, finite elements and spectral methods work well for solving 
partial (integro-)differential equations P(I)DEs in low dimensions, they suffer from the curse of dimensionality
and, as a consequence, become infeasible in high dimensions. 
Recently, different deep learning based approaches for solving high-dimensional PDEs have been proposed \citep{raissi2017physics, raissi2019physics, han2017deep,han2018solving, sirignano2018dgm, berg2018unified, beck2021deep, lu2021deepxde, bruna2024neural}. They can directly be used to solve continuous-time stochastic control problems that admit an explicit solution for the optimal control in terms of the value function since in this case, the expression for the optimal control can be plugged into the HJB equation, which then reduces to a parabolic PDE. 
On the other hand, if the optimal control is not available in closed form, it cannot be plugged into the HJB equation, but instead, has to be approximated numerically while at the same time solving a parabolic PDE. Such implicit optimal control problems can no longer be solved directly with one of the deep learning methods mentioned above but require a specifically designed iterative approximation procedure. For low-dimensional problems with non-explicit optimal controls, a standard approach from the reinforcement learning (RL) literature is to use generalized policy iteration (GPI), a class of iterative algorithms that simultaneously approximate the value function and optimal control \citep{jacka2017policy, sutton2018reinforcement}. Particularly popular are actor-critic methods going back to \citet{werbos1992approximate}, which have a separate memory structure to  represent the optimal control independently of the value function. However, classical GPI schemes become impractical in high dimensions as the PDE and optimization problem both have to be discretized and solved for every point in a finite grid. This raises the need for meshfree methods to solve implicit continuous-time stochastic control problems in high dimensions with continuous action space. Several local approaches based on a time-discretization of \eqref{sde} have been explored by e.g. \cite{han2016deep}, \cite{nusken2021solving}, \cite{hure2021deep}, \cite{bachouch2022deep}, \cite{ji2022solving}, \cite{li2024neural}, 
\cite{domingo2024adjoint}, \cite{domingo2024stochastic}. Alternatively,  (deep) RL methods can be used,
for instance,  Q-learning type algorithms such as DQN \citep{mnih2013playing} or C51 \citep{bellemare2017distributional}, see also \cite{wang2020reinforcement}, \cite{jia2023q}, \cite{gao2024reinforcement}, \cite{szp}; or  actor critic approaches such as DDPG \citep{lillicrap2015continuous}, SAC \citep{haarnoja2018soft},
A2C/A3C \citep{mnih2016asynchronous}, PPO \citep{schulman2017proximal} or TRPO  \citep{schulman2015trust}.
However, these RL algorithms are model-free, that is, they do not explicitly take into account the underlying dynamics \eqref{sde} of the control problem \eqref{pr} but instead, solely rely on sampling from the environment. 
As a result, they are less accurate in cases where the system dynamics are known (see Figure \ref{fig3} below). 
On the other hand, the difficulty of solving high-dimensional PDEs is further exacerbated if the 
system dynamics includes stochastic jumps since in this case, the 
HJB equation \eqref{HJB} requires the computation of the jump expectations 
$\E \, V (t, x + \gamma(t,x,Z_1,a))$ for every space-time point $(t,x)$ sampled from the domain.


In this paper, we introduce a deep model-based approach for stochastic control problems with 
jumps that takes the system dynamics \eqref{sde} into account by leveraging the HJB equation \eqref{HJB}. 
This removes the need to simulate the underlying jump-diffusion \eqref{sde} and, as a result, 
avoids discretization errors. Our approach combines GPI with a PIDE solving method.
It approximates the value function and optimal control in an actor-critic fashion with two neural networks trained  iteratively on sampled data from the space-time domain. As such, it has the advantage that it provides global approximations of the value function and optimal control available for all space-time points, which can be evaluated rapidly in online applications. We develop two related algorithms. The first one, GPI-PINN, 
approximates the value function by training a neural network to minimize the residuals of the HJB equation 
\eqref{HJB}, following a physics-inspired neural network (PINN) approach 
\citep{raissi2017physics, raissi2019physics} while leveraging Proposition~\ref{Prop1} below to avoid the direct computation of the gradient $\nabla_x V$ and Hessian $\nabla_x^2 V$
in the Hamiltonian \eqref{Ham}. GPI-PINN can be viewed as an extension of the method proposed by \citet{duarte2024machine} adapted to a finite-horizon setup with time-dependence and a terminal condition 
in the HJB equation, leading to time-dependent optimal control strategies and value function, see  also \cite{dupret2024deep}. 
It works well in high dimensions for control problems without jumps in the underlying dynamics
\eqref{sde} $(\gamma = 0)$, but becomes inefficient in the presence of jumps as it requires 
the computation of the jump-expectation $\E \, V (t, x + \gamma(t,x,Z_1,a))$ at numerous sample 
points $(t,x)$ in every iteration of the algorithm. To address this, our second algorithm, GPI-CBU, relies on a continuous-time Bellman updating rule to approximate the value function, thereby circumventing the computation of gradients, Hessians and jump-expectations altogether. This makes it highly efficient for high-dimensional stochastic control problems with jumps, even when the control is not available in closed form. 

We illustrate the accuracy and scalability of 
our approach in different numerical examples and provide comparisons with popular 
RL and deep-learning control methods. Proofs of theoretical results and additional numerical 
experiments are given in the Appendix. \vspace{-2mm}
\section{General approach} 
\label{Section2}

Let $\alpha \colon [0,T) \times D \to A$ be a feedback control such that equation 
\eqref{sde} has a unique solution $X^{\alpha}$ and consider the corresponding value function
\[
V^{\alpha}(t,x) =   \E \edg{\int_t^T f(s, X^\alpha_s, \alpha(s, X^{\alpha}_{s-})) ds + F(X^\alpha_T) \, \Big| \,
	X^\alpha_t = x}, \quad (t, x) \in [0, T] \times D.
\]
Under appropriate assumptions, one obtains the following two results from 
standard arguments\footnote{see e.g. the arguments in the proofs of Theorems 1.3.1 and 2.2.4 
	in \cite{bouchard2021}.}.

\begin{theorem}[Feynman--Kac Formula] \label{thm:FK}
	$V^{\alpha}$ satisfies the PIDE
	\[
	\partial_t V^{\alpha}(t, x) + H(t, x, V^{\alpha}, \alpha(t,x))=0, \quad V^{\alpha}(T,x) = F(x) .
	\]
\end{theorem}

\begin{theorem}[Verification Theorem] \label{thm:ver}
	Let $v \in \mathcal{V} \cap C([0,T] \times D)$ be a solution of the HJB equation \eqref{HJB}
	such that there exists a measurable mapping $\hat{\alpha}: [0,T) \times D \to A$ satisfying 
	\[
	\hat{\alpha}(t,x) \in \argmax_{a \in A} H(t, x,  v, a) \ \quad \text{for all } (t,x) \in [0,T) \times D \label{theorem1} 
	\]
and the controlled jump-diffusion equation \eqref{sde} admits a unique solution for each initial condition $x \in D$. Then $v = V$ and $\hat{\alpha}$ is an optimal control.
	\label{Th2} \vspace{-1mm}
\end{theorem}

Based on Theorems \ref{thm:FK} and \ref{thm:ver}, we iteratively approximate the value function 
$V$ and optimal control $\alpha^*$ with neural networks\footnote{Using a $C^2$-activation function
	in the network $V_{\theta}$ ensures that it belongs to $C^{2}([0,T] \times \R^d)$.} 
$V_{\theta} \colon [0,T] \times D \to \R$ 
and $\alpha_\phi \colon [0,T] \times D \to A$.
For given $\alpha_{\phi}$, we train $V_{\theta}$ so as
to solve the controlled HJB equation
\begin{equation} \label{HJBc}
	\partial_t V_\theta(t, x) + H(t, x, V_\theta, \alpha_{\phi}(t,x))=0, \quad V_\theta(T,x) = F(x) \, ,
\end{equation}
while for given $V_{\theta}$, $\alpha_{\phi}$ is trained with the goal to maximize the Hamiltonian
$H(t, x, V_{\theta}, \alpha_{\phi}(t,x)).$

In the following, we introduce two different training objectives for updating the value network $V_\theta$,
leading respectively to the algorithms GPI-PINN and GPI-CBU. GPI-PINN uses a PINN-type loss together 
with a trick adapted from \cite{duarte2024machine} to bypass the explicit computation of 
gradients and Hessians, whereas GPI-CBU relies on a continuous-time Bellman updating rule with an 
expectation-free version of the Hamiltonian, thereby also avoiding the computation of the jump-expectations
in \eqref{Ham}.

\section{GPI-PINN} 
\label{sec:pinn}

GPI-PINN, described in Algorithm \ref{Algo1} below, relies on a PINN approach to minimize the 
residuals of the controlled HJB equation \eqref{HJBc} in the value function approximation step.
To avoid explicit computations of the gradient $\nabla_x V_\theta(t,x)$ and 
Hessian $\nabla^2_x V_\theta(t,x)$, which appear in the Hamiltonian \eqref{Ham}, 
we use the following trick, adapted from \citet{duarte2024machine}.
\begin{proposition}\label{Prop1}
	Consider a function $v \in \mathcal{V}$ together with a pair $(t,x) \in [0,T) \times D$. Define the function 
	$\psi: \mathbb{R} \to \mathbb{R}$ \vspace{-2mm} by
	\begin{equation*}
		\psi(h) := \sum_{i=1}^{n} v \hspace{-0.3mm}
		\left(t + \frac{h^2}{2n}, x+\frac{h}{\sqrt{2}}\sigma_i(t,x,a)+\frac{h^2}{2n}\beta(t,x,a)\right) , \vspace{-1mm}
	\end{equation*} where $\sigma_i(t,x,a)$ is the $i^{th}$ column of the $d\times n$ matrix $\sigma(t,x,a)$. Then,
	\begin{align*}
		\psi''(0) =  \partial_t v(t,x) &+ \beta^\top\hspace{-0.5mm}(t, x, a)  \, \nabla_xv(t,x) + \frac{1}{2} \normalfont \text{Tr} \left[ \sigma \sigma^\top\hspace{-0.5mm}(t,x,a) \nabla^2_x v(t,x) \right] .
		\label{eqprop}
	\end{align*}   \vspace{-4mm}
\end{proposition} 
Proposition \ref{Prop1} makes it possible to replace the computation of gradients and Hessians of 
$v$ by evaluating the univariate function $\psi''(0)$, the cost of which, using automatic differentiation, is a small multiple 
of $n \cdot cost(v)$.

To formulate GPI-PINN, we need the extended Hamiltonian
\begin{equation} \label{extHam}
	\mathcal{H}(t,x,v,a) := \partial_t v(t,x) + H(t,x,v,a),
\end{equation}
which by Proposition \ref{Prop1}, can be written as 
\begin{equation*}
	\mathcal{H}(t,x,v,a) =  \psi''(0) + f(t,x,a) + \lambda(t,x,a) \mathbb{E} \big[  v(t,x+\gamma(t,x,Z_1,a)) - v(t,x) \big] .
\end{equation*}	
In Algorithm \ref{Algo1}, we simplify the notation by using
$\mathcal{H}(t,x, \theta, \phi) := \mathcal{H}(t,x, V_{\theta}, \alpha_{\phi}(t,x)).$

\begin{algorithm}[h]
\caption{GPI-PINN}
\label{Algo1}
\begin{algorithmic}
	\STATE Initialize admissible weights $\theta^{(0)}$ for $V_{\theta}$ and $\phi^{(0)}$ for  
	$\alpha_{\phi}$. Choose proportionality factors $\xi_1,\xi_2 > 0$ and set epoch $k=0$. 
		\REPEAT \vspace{0.5mm}
		\STATE  \textbf{Step 1:} Update the value network $V_{\theta^{(k+1)}}$ for a given control  
		$\alpha_{\phi^{(k)}}$ by minimizing the loss
		\vspace{-0.5mm}
	\begin{equation*}
		\theta^{(k+1)} = \argmin_{\theta} \mathscr{L}_1(\theta, \phi^{(k)}) \, , \vspace{-3mm} \label{theta1}
		\end{equation*}
		where \vspace{-1.5 mm}
	\begin{align}
	\hspace{-2mm} \mathscr{L}_1( \theta,\phi ) &=  \xi_1	\, \mathbb{E}_{(t,x)\sim \mu}\mathcal{H}^2(t,x,\theta,\phi)  + \xi_2 \, \mathbb{E}_{x \sim \nu} \left(V_\theta(T,x) - F(x)\right)^2 \label{squareloss}
		\end{align}
		\vspace{-1.5mm}
		\STATE \textbf{Step 2:} Update the control network $\alpha_{\phi^{(k+1)}}$ for a given value network $V_{\theta^{(k+1)}}$ by minimizing the loss
	\begin{equation*}
	\phi^{(k+1)} = \argmin_{\phi} \mathscr{L}_2(\theta^{(k+1)}, \phi) \, ,\vspace{-3mm} \label{theta2b}
		\end{equation*}
		where \vspace{-1.75mm}
	\begin{align}
	\mathscr{L}_2(\theta, \phi) & = - \mathbb{E}_{(t,x) \sim \mu} \mathcal{H}\big(t,x,\theta, \phi \big)  
	\label{squareloss2} 
		\end{align} \vspace{-2mm}
		\STATE $k \leftarrow k+1$
		\UNTIL{ some convergence criterion is satisfied. 
		}
		\STATE  \textbf{return} \hspace{0.4mm}$V_{\theta^{(k)}}$ and 
		$\alpha_{\phi^{(k)}}$ and set  $k_* \leftarrow k$.
	\end{algorithmic}  	
\end{algorithm}

The loss function $\mathscr{L}_1$ in \eqref{squareloss} consists of two terms. The first
represents the expected PIDE residual in the interior of the space-time domain with respect to a suitable 
measure $\mu$ on $[0,T) \times D$. The second term penalizes violations of the 
terminal condition according to a measure $\nu$ on $D$. Hence, $\mathscr{L}_1$ 
measures how well the function $V_\theta$ satisfies the controlled HJB equation \eqref{HJBc} 
corresponding to a control $\alpha_{\phi}$. In every epoch $k$, 
the goal in Step 1 is therefore to find a parameter vector $\theta$ such that the value network $V_\theta$ 
minimizes the error $\mathscr{L}_1(\theta, \phi^{(k)})$. We do this with a mini-batch stochastic 
gradient method which updates the measures $\mu$ and $\nu$ according to the residual-based 
adaptive distribution (RAD) method of \cite{wu2023comprehensive}, as it is known to significantly improve 
the accuracy of PINNs. In Step 2, we minimize $\mathscr{L}_2(\theta^{(k+1)}, \phi)$ with 
respect to $ \phi$. This corresponds to choosing the control $\alpha_\phi$ so as to maximize 
the extended Hamiltonian $\cal{H}$ (or equivalently $H$); see \cite{duan2023relaxed}, \cite{dupret2024deep} and \cite{cohen2025neural} for theoretical convergence results supporting this approach.

Since the expectations in $\mathscr{L}_1$, $\mathscr{L}_2$ and the 
extended Hamiltonian \eqref{extHam} are typically not available in closed form, we replace them by 
sample-based estimates. First, we estimate $\mathcal{H}(t,x,\theta, \phi)$ with 
$\widehat{\mathcal{H}}(t,x,\theta, \phi)$ by approximating the jump-expectation
\begin{equation} \label{jumpappr}
\E \, V_{\theta}(t,x+\gamma(t,x,Z_1,a_{\phi}(t,x))) 
\approx \frac{1}{J} \sum_{j=1}^J V_{\theta}(t,x+\gamma(t,x,z_j,a_{\phi}(t,x)))
\end{equation}
for points $(z_j)_{j=1}^J$ in $E$ sampled from the distribution $\mathcal{Z}$ of $Z_1$. Then, 
we approximate $\mathscr{L}_1$ and $\mathscr{L}_2$ in every time step by 
\begin{equation*} \widehat{\mathscr{L}}_1(\theta,\phi^{(k)}) = \frac{\xi_1}{M_1} \sum_{m=1}^{M_1}  \Big( \widehat{\mathcal{H}}\big(t_m,x_m,\theta,\phi^{(k)}\big)\Big)^2 + \frac{\xi_2}{M_2}\sum_{m=1}^{M_2} \left( V_\theta(T, y_m) - F(y_m) \right)^2  \label{How1}  \end{equation*}
and \vspace{-1mm}
\begin{equation*} 	\widehat{\mathscr{L}}_2(\theta^{(k+1)}, \phi)=   -\frac{1}{M_1} \sum_{m=1}^{M_1}  \widehat{\mathcal{H}}\hspace{-0.5mm}\left(t_m,x_m,\theta^{(k+1)},  \phi\right) ,  \  \label{How2} 
\end{equation*} 
where $(t_m, x_m)_{m=1}^{M_1} \in [0, T) \times D$ and $(y_m)_{m=1}^{M_2} \in D$ are sampled from $\mu$ and $\nu$, respectively.
In every epoch $k = 0 , \dots, k_*-1$, we initialize $\theta^{(k)}_0 := \theta^{(k)}$ and make 
$N_1$ gradient steps \vspace{1mm}
{\small 
\begin{align}
\theta^{(k)}_{i+1} &= \theta^{(k)}_{i}  - \eta_1 \nabla_\theta \widehat{\mathscr{L}}_1(\theta^{(k)}_i,\phi^{(k)}) = \theta^{(k)}_{i} 
-  \frac{2 \eta_1 \xi_1}{M_1}\sum_{m=1}^{M_1}  \widehat{\mathcal{H}}\big(t_m,x_m,\theta^{(k)}_i,\phi^{(k)}\big) \nabla_\theta \widehat{\mathcal{H}}\big(t_m,x_m,\theta^{(k)}_i,\phi^{(k)}\big)\nonumber \\[-0.65mm]
&\hspace{44.5mm}- \frac{2 \eta_1 \xi_2}{M_2} \sum_{m=1}^{M_2} \left(   V_{\theta^{(k)}_i}\hspace{-0.4mm}(T, y_m) - F(y_m) \right) \nabla_\theta V_{\theta^{(k)}_i}\hspace{-0.4mm}(T, y_m ), \label{grad1}
\end{align}}
$i = 0, \dots, N_1-1$, to obtain $\theta^{(k+1)} = \theta^{(k)}_{N_1}$. 
Then, we initialize $\phi^{(k)}_0 = \phi^{(k)}$ and perform $N_2$ gradient steps
\begin{equation} \label{grad2}
\begin{aligned}
\phi^{(k)}_{i+1} &= \phi^{(k)}_i  - \eta_2  \nabla_\phi \widehat{\mathscr{L}}_2(\theta^{(k+1)},\phi^{(k)}_i) 
= \phi^{(k)}_i +  \frac{\eta_2}{M_1} \sum_{m=1}^{M_1}  
\nabla_\phi\widehat{\mathcal{H}}\hspace{-0.5mm}\left(t_m,x_m,\theta^{(k+1)},  \phi^{(k)}_i\right),
\end{aligned}
\end{equation} $i = 0, \dots, N_2 -1,$
to get $\phi^{(k+1)} = \phi^{(k)}_{N_2}$.

GPI-PINN yields global approximations $V_{\theta^{(k_{*})}}$ and
$\alpha_{\phi^{(k_{*})}}$ of the value function and optimal control on the whole space-time domain $[0,T] \times D$.
By using Proposition \ref{Prop1}, it avoids the computation of the gradients and Hessians
appearing in the Hamiltonian \eqref{Ham}. However, it still has two drawbacks that make it inefficient for 
high-dimensional control problems with jumps. First, it has to approximate the 
jump-expectations $\mathbb{E}\big[V_\theta(t_m,x_m+\gamma(t_m,x_m,Z_1,\alpha_{\phi}(t_m,x_m)))\big]$ 
for all sample points $(t_m,x_m)$, $m = 1, \dots, M_1$, in each of the gradient steps \eqref{grad1}--\eqref{grad2}
and all epochs $k = 0, 1, \dots, k_*$; see \eqref{jumpappr}.
Secondly, since the Hamiltonian is already a second-order integro-differential operator, 
the gradient steps  $\nabla_\theta \widehat{\mathcal{H}}$ in \eqref{grad1} require the computation of third order derivatives, which is numerically costly. 

\section{GPI-CBU} 
\label{sec:cbu}

GPI-CBU addresses the shortcomings of GPI-PINN by using a value function updating rule 
based on the expectation-free operator
$G_{\zeta} : [0,T] \times D \times E \times \mathcal{V} \times A \to \mathbb{R}$ given by
\begin{equation*} \label{Ht} 
\begin{aligned}
G_{\zeta}(t,x, z, v, a) := v(t,x)+ \zeta & \big[ \partial_t v(t,x) + f(t,x,a)    +	\beta^\top\hspace{-0.5mm} (t, x, a) \, \nabla_x v(t, x) \\
	 &  \hspace{-28mm}+  \frac{1}{2} \normalfont \text{Tr} \hspace{-0.3mm}\left[ \sigma \sigma^\top\hspace{-0.5mm}(t,x,a)\nabla^2_x v(t,x) \right]   + \lambda(t,x,a)  \left( v(t,x+\gamma(t,x,z,a)) - v(t,x) \right) \big] 
	\end{aligned}
\end{equation*}
for a scaling factor $\zeta \in \R$. 

\begin{proposition} 	\label{Prop2}
Let $X^\alpha$ be a solution of the jump-diffusion equation \eqref{sde} corresponding to a 
feedback control $\alpha \colon [0,T) \times D \to A$ with associated value function 
$V^{\alpha} \in C^{1,2}([0,T] \times D)$. For given $t \in [0,T)$, let $Y_t$ be a $D$-valued 
random variable independent of $Z_1$ such that $\E \, G_{\zeta}^2 (t,Y_t, Z_1, V^{\alpha}, \alpha(t,Y_t)) 
< \infty$. Then $V^\alpha(t,Y_t) = g(Y_t)$ for the Borel measurable function $g :  D \to \R$ minimizing 
the mean squared error
\[
\mathbb{E} \edg{ \big( g(Y_t) - G_{\zeta}(t,Y_t, Z_1, V^{\alpha}, \alpha(t, Y_t)) \big)^2 }.
\]
\end{proposition}

Proposition \ref{Prop2} suggests to update\footnote{By \eqref{Bell}, the update rule \eqref{notyet} 
corresponds to $V_{\theta^{(k+1)}} \hspace{-0.5mm}= \hspace{-0.5mm}T^\alpha V_{\theta^{(k)}} \hspace{-1.2mm}:= \hspace{-0.7mm}V_{\theta^{(k)}}\hspace{-0.3mm}+\hspace{-0.3mm} \zeta \mathcal{H}(\cdot,V_{\theta^{(k)}}, \alpha(\cdot))$ for the continuous-time Bellman updating (CBU) operator $T^\alpha$ 
associated with the feedback control $\alpha$ and fixed point $T^\alpha V^\alpha = V^\alpha$.}  the value function parameters according to
\begin{equation}
\theta^{(k+1)}  = \argmin_{\theta}  \mathbb{E} \int_0^T \Big(  V_\theta(t, Y_t)  - 
G_{\zeta}\hspace{-0.1mm}\big(t, Y_t , Z_1,  V_{\theta^{(k)}}, \alpha(t, Y_t)\big) \Big)^2 dt . \label{notyet}  
\end{equation} 
By adding a penalty term enforcing the terminal condition, we obtain the recursive scheme
\begin{equation} \label{Belupdate}
\theta^{(k+1)} = \argmin_\theta \mathscr{L}^{(k)}_1(\theta) \vspace{-1mm}
\end{equation} for the loss
\begin{align*}
		\mathscr{L}^{(k)}_1(\theta) \hspace{-0.4mm} &= \hspace{-0.4mm} \xi_1 \mathbb{E}_{(t,x,z)\sim \mu \otimes \mathcal{Z}}  \left(V_\theta(t,x) - G_{\zeta}(t,x,z,\theta^{(k)},\phi^{(k)}) \right)^2  + \xi_2 \mathbb{E}_{ x \sim \nu}  \left(V_\theta(T,x) - F(x) \right)^2, 
\end{align*} 
where we use the notation $
G_{\zeta}(t,x,z,\theta, \phi) := G_{\zeta}(t,x,z,V_\theta, \alpha_\phi(t,x))$.
To implement \eqref{Belupdate}, we approximate $\mathscr{L}^{(k)}_1(\theta)$ with
{\small\begin{equation}	\hspace{-2mm} \widehat{\mathscr{L}}^{(k)}_1\hspace{-0.4mm}(\theta) = \frac{\xi_1}{M_1} \hspace{-0.2mm} \sum_{m=1}^{M_1}  \hspace{-0.5mm}  \left( \hspace{-0.2mm}V_\theta(t_m, x_m) \hspace{-0.2mm}  - \hspace{-0.2mm} G_{\zeta}\big(t_m,x_m,z_m,\theta^{(k)},\phi^{(k)}\big) \hspace{-0.3mm} \right)^2 \hspace{-0.3mm} + \hspace{-0.4mm}  \frac{\xi_2}{M_2} \hspace{-0.4mm} \sum_{m=1}^{M_2} \hspace{-0.6mm} \left( V_\theta(T, y_m) - F(y_m) \right)^2 \hspace{-1mm} \hspace{-1mm} \label{How3}
\end{equation}}for $(t_m, x_m, z_m)_{m=1}^{M_1} \in [0,T) \times D \times E$ sampled from 
$\mu \otimes \mathcal{Z}$ and $(y_m)_{m=1}^{M_2} \in D$ sampled from $\nu$.
In every epoch $k=0,\ldots,k_*-1$,  we initialize $\theta^{(k)}_0 = \theta^{(k)}$ and perform $N_1$ gradient steps
{\small \begin{equation} 	\label{grad3} 
	\begin{aligned}
	\theta^{(k)}_{i+1}  \hspace{-0.25mm}&=  \hspace{-0.25mm}	\theta^{(k)}_i \hspace{-0.5mm}- \eta_1 \nabla_\theta \widehat{\mathscr{L}}^{(k)}_1\hspace{-0.3mm}(\theta^{(k)}_i) = \theta^{(k)}_i -  \frac{2 \eta_1 \xi_2}{M_2}\sum_{m=1}^{M_2} \left( V_{\theta^{(k)}_i}\hspace{-0.4mm}(T, y_m) - F(y_m) \right) \nabla_\theta V_{\theta^{(k)}_i}\hspace{-0.4mm}(T, y_m) \\[-1.5mm]
&\hspace{18mm}-  \frac{2 \eta_1 \xi_1}{M_1} \sum_{m=1}^{M_1}  \hspace{-0.43mm} \left( V_{\theta^{(k)}_i}\hspace{-0.4mm}(t_m, x_m) - G_{\zeta}\big(t_m,x_m,z_m,\theta^{(k)}_0,\phi^{(k)}\big) \right) \hspace{-0.5mm} \nabla_\theta V_{\theta^{(k)}_i}\hspace{-0.4mm}(t_m, x_m)  \, , \hspace{-2mm} 
\vspace{-1.5mm}
\end{aligned} \end{equation}}$i=0,...,N_1-1$, to obtain $\theta^{(k+1)} = \theta^{(k)}_{N_1}$. 
To update the control network $\alpha_{\phi}$, we introduce the operator
{\small\begin{align}
	\mathcal{G}(t,x,z,\theta,\phi) &:= \partial_t V_\theta(t,x) + f(t,x,\alpha_\phi(t,x))    +	\beta^\top\hspace{-0.5mm} (t, x, \alpha_\phi(t,x)) \, \nabla_x V_\theta(t, x) \nonumber
	\\[-0.45mm]
	  &\hspace{-2cm}+  \frac{1}{2} \normalfont \text{Tr} \hspace{-0.3mm}\left[ \sigma \sigma^\top\hspace{-0.5mm}(t,x,\alpha_\phi(t,x))\nabla^2_x V_\theta(t,x) \right]   + \lambda(t,x,\alpha_\phi(t,x))  \left( V_\theta(t,x+\gamma(t,x,z,\alpha_\phi(t,x))) - V_\theta(t,x) \right) ,  \nonumber
\end{align}}which is an expectation-free version of the extended Hamiltonian $\mathcal{H}$ that,
instead of the jump-expectation $\E \, V_{\theta}(t,x+\gamma(t,x,Z_1,a_{\phi}(t,x)))$,
only contains a single jump $V_{\theta}(t,x+\gamma(t,x,z,a_{\phi}(t,x)))$, $z\in E$.\
GPI-CBU updates the parameters of the control network $\alpha_{\phi}$ according to
\[
\phi^{(k+1)} = \argmin_\phi \mathscr{L}^{(k+1)}_2(\phi)
\] for the loss  \vspace{1mm}
\begin{equation*}
	\mathscr{L}^{(k+1)}_2(\phi) = - \mathbb{E}_{(t,x,z)\sim \mu \otimes \mathcal{Z}}   \ \mathcal{G}(t,x,z,\theta^{(k+1)},\phi)   = -\mathbb{E}_{(t,x)\sim \mu}   \mathcal{H}(t,x,\theta^{(k+1)},\phi). \vspace{1mm} \label{key3}
\end{equation*}
We hence minimize the sample estimate
\begin{equation} 
\widehat{\mathscr{L}}^{(k+1)}_2(\phi) =  - \frac{1}{M_1} \sum_{m=1}^{M_1} \mathcal{G}(t_m,x_m, z_m,\theta^{(k+1)},\phi) \, , \label{loss4} \vspace{-1mm}
\end{equation}
by setting $\phi^{(k)}_0 = \phi^{(k)}$ and making $N_2$ gradient steps  
\begin{equation} \label{grad4}  
\begin{aligned} 
\phi^{(k)}_{i+1} &= \phi^{(k)}_i  - \eta_2 \nabla_\phi \widehat{\mathscr{L}}^{(k+1)}_2(\phi^{(k)}_i) = \phi^{(k)}_i + \frac{\eta_2}{M_1} \sum_{m=1}^{M_1} \nabla_\phi \mathcal{G}(t_m,x_m, z_m,\theta^{(k+1)}, \phi^{(k)}_i),
\end{aligned} \vspace{-0.25mm}
\end{equation} $i=0,\ldots,N_2-1$, to obtain $\phi^{(k+1)}=\phi^{(k)}_{N_2}$.  \vspace{-1.2mm}

 \begin{algorithm}[h]
	\caption{GPI-CBU}
	\label{Algo2}
	\begin{algorithmic}
		\STATE Initialize admissible neural weights  $\theta^{(0)}$ for $V_{\theta}$ and $\phi^{(0)}$ for $\alpha_{\phi}$. Choose learning rates $\eta_1, \eta_2$, proportionality factors $\xi_1,\xi_2$
		and numbers of gradient steps $N_1, N_2$. Set epoch $k=0$.
		
		\REPEAT
		
		\STATE 
		\textbf{Step 0:} Generate $M_1$ sample points $(t_m, x_m, z_m) \in [0,T) \times D \times E$ from $\mu \otimes \mathcal{Z}$ and $M_2$ sample points $y_m \in D$ from $\nu$. \vspace{0.5mm}
		\STATE \textbf{Step 1:} Update $V_{\theta^{(k+1)}}$ by minimizing the loss \eqref{How3}
		\begin{equation*}
		\theta^{(k+1)} = \argmin_{\theta}\widehat{\mathscr{L}}^{(k)}_1(\theta) \, , \vspace{-1mm} \label{theta1c}
		\end{equation*}
		with $N_1$ gradient steps \eqref{grad3}.
		
		\STATE \textbf{Step 2:} Update $\alpha_{\phi^{(k+1)}}$ by minimizing the loss \eqref{loss4}
		\begin{equation*}
			\phi^{(k+1)} = \argmin_{\phi}\widehat{\mathscr{L}}^{(k+1)}_2(\phi) \, , \vspace{-1mm} \label{theta1cc}
		\end{equation*}
		with $N_2$ gradient steps \eqref{grad4}.
		\vspace{0.75mm}
		\STATE $k \leftarrow k+1$
		\UNTIL{some convergence criterion is satisfied. 
		}
		\STATE \textbf{return} $V_{\theta^{(k)}}$ and $\alpha_{\phi^{(k)}}$ and set $k_* \leftarrow k$.
	\end{algorithmic} 
\end{algorithm}  

Like GPI-PINN, GPI-CBU leverages Proposition \ref{Prop1} to bypass the computation of the gradients 
$\nabla_x V_{\theta}$ and Hessians $\nabla^2_x V_{\theta}$. In addition, 
it avoids the costly computation of the jump-expectations \eqref{jumpappr} of $V_\theta$ at each sample point $(t_m,x_m)$. Also, it 
it does not have to compute third-order derivatives like \eqref{grad1} when updating the value network 
since only $\nabla_\theta V_\theta$ is needed in \eqref{grad3}. This is a consequence of the fact that
the value update rule \eqref{grad3} of GPI-CBU is recursive as $G_\zeta$ depends on the value weights 
$\theta^{(k)}$ computed in the previous epoch in contrast to the residual-based GPI-PINN.
On the other hand, since GPI-PINN averages over different jumps in each update, 
it exhibits more stable convergence than GPI-CBU; see Figure \ref{fig1} below.  
The proportionality factors $\xi_1, \xi_2 $ and the scaling factor $\zeta$ are hyperparameters.
$\xi_1$ and $\xi_2$ can be fine-tuned following \cite{wang2022and}. While Proposition \ref{Prop2}
holds for any scaling factor $\zeta \in \R$, its choice affects the numerical performance of GPI-CBU.
In the numerical experiments of this paper, we set $\zeta = 1$ as it provides a good trade-off between convergence speed and accuracy of the improvements in GPI-CBU. On the other hand, negative scaling factors 
failed to converge to the true solution in all our experiments with exploding losses 
$\widehat{\mathscr{L}}_1^{(k)}$ and $\widehat{\mathscr{L}}_2^{(k)}$ after only a few epochs. Alternatively, one could consider an adaptive scaling factor $\zeta_k > 0$ depending on the epoch $k$. More details on hyperparameter fine-tuning are given in Appendix \ref{dgm}. Finally, we note that the policy updating rule of GPI-CBU \eqref{grad4} is equivalent to that of GPI-PINN \eqref{grad2}, except that it circumvents the computation of the jump-expectations in 
$\widehat{\cal H}$ since it uses the expectation-free operator $\mathcal{G}$.

\section{Numerical experiments} 
\label{sec:num} 
In our numerical experiments, we use the Deep Galerkin Method (DGM) architecture of \citet{sirignano2018dgm} for both the value and  optimal control networks as it has been shown to empirically improve PINN performance. Details of the network design and hyperparameters are given in Appendix \ref{dgm}.
\subsection{Linear-quadratic regulator with  jumps} 
\label{Sec:LQR}
We first consider the linear-quadratic regulator (LQR) problem with jumps \vspace{-0.5mm}
 \begin{equation} \label{LQR} 
\inf_{\alpha} \mathbb{E}\left[ \int_0^T c_1 \|\alpha_s\|^2_2 \, ds + c_2 \|X^\alpha_T \|^2_2  \right] , 
\vspace{-1mm}
\end{equation}
where the infimum is over $d$-dimensional predictable processes $(\alpha_t)_{0 \leq t \leq T}$ and $X^\alpha$ is a $d$-dimensional process with controlled dynamics \vspace{-1mm}
 \begin{equation} \label{Xa}
 dX^\alpha_t = \alpha_t  dt + B  dW_t + d \sum_{j=1}^{M^\alpha_t} Z_j \,  , \ \  X_0 = x \in \mathbb{R}^d  
 \vspace{-1mm}
\end{equation}
for a $d\times d$ matrix $B$, a $d$-dimensional Brownian motion $W$, a Cox process $M^\alpha$ with intensity $\lambda^\alpha_t = \Lambda_1 + \Lambda_2 \|\alpha_t\|^2_2\,$ for constants
$\Lambda_1, \Lambda_2 \geq 0$ and independent i.i.d.\ zero-mean square integrable $d$-dimensional random vectors $Z_j$ with $\mathbb{E}[\|Z_j\|^2_2] = \upsilon$, $\, j=1,2,\dots$ 

Problem \eqref{LQR} is of the general form \eqref{pr} if instead of the infimum, one considers the supremum of the negative of the expectation. It can be reduced to a one-dimensional ODE which has a 
closed form solution in the special case $\Lambda_2 = 0$ and admits a very precise numerical
solution via a standard ODE solver, such as Runge--Kutta, if $\Lambda_2 > 0$ (see Appendix \ref{lqr2}). This provides 
highly accurate reference solutions $V$ and $\alpha^*$ for the value function and optimal control.

\begin{figure}[h]
	\centering
	\subfigure{ \hspace{-1mm}
		\includegraphics[width=0.485\columnwidth, height=4.2cm]{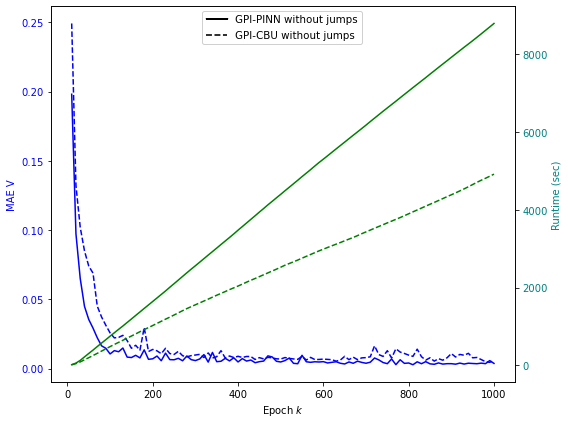}
		\label{fig:sub44}
	}
	 \hspace{-3mm}
	\subfigure{
	\includegraphics[width=0.49\columnwidth, height=4.2cm]{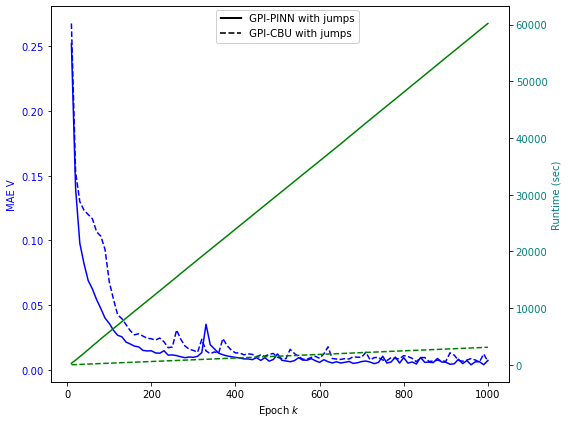}
	\label{fig:sub33}
	} \vspace{-1.95mm}
	\caption{{\small Comparison of ${\rm MAE}_V$ (blue) and runtime in seconds (green) of GPI-PINN (solid lines) and  GPI-CBU (dashed lines) for a 10-dimensional LQR problem \eqref{LQR}
	without jumps (left) and with jumps (right).}} \vspace{-4mm}
	\label{fig1} 
\end{figure}

Figure \ref{fig1} compares the performances of GPI-PINN and GPI-CBU on a 10-dimensional $(d=10)$
LQR problem of the form \eqref{LQR} with $T = 1$, $B = I_d$, $c_1 = 1$, $c_2 = 1/4$
without jumps ($\Lambda_1 = \Lambda_2 = 0$) and with jumps ($\Lambda_1 = 0$, $\Lambda_2 = 2$
and $d$-dimensional standard normal jumps $Z_j$).
It shows mean absolute errors of $V_{\theta^{(k)}}$ with respect to $V$ 
given by ${\rm MAE}_V = \frac{1}{M} \sum_{i=1}^{M} |V_{\theta^{(k)}}\hspace{-0.37mm}(t_i, x_i) - V(t_i, x_i) | $ 
on a test set of size $M$ uniformly sampled from $[0,1] \times [-2.5,2.5]^d$ together with runtimes
as functions of the number of epochs $k$.
It can be seen that GPI-PINN and GPI-CBU exhibit similar convergence in the number of epochs.
The residual-based approach of GPI-PINN makes it more stable than GPI-CBU; see also 
\citet{baird1995residual}. On the other hand, GPI-CBU has a lower computational cost. 
This is already the case without jumps (left plot of Figure \ref{fig1}) since it avoids the numerical 
evaluation of third-order derivatives but becomes much more significant with jumps 
(right plot of Figure \ref{fig1}) as it also circumvents the numerical integration of the jumps.\vspace{-2mm}

\begin{figure}[h]
	\centering
	\begin{minipage}{.5\textwidth}
		\includegraphics[width=0.95\columnwidth, height=4.2cm]{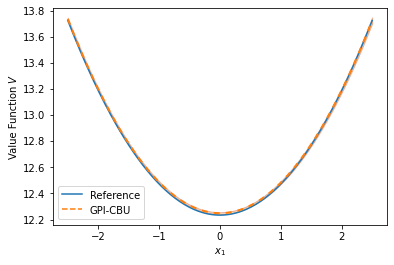}
		\label{fig:test10} \vspace{6mm}
	\end{minipage} \hspace{-4mm}
	\begin{minipage}{.5\textwidth} 	\vspace{-5mm}
		\includegraphics[width=0.96\columnwidth, height=4.2cm]{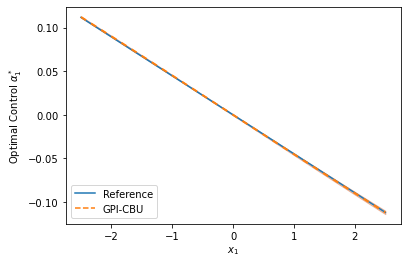}
		\label{fig:test20}
	\end{minipage}\vspace*{-6mm}
\caption{\small Value function $V(0,x)$ (left) and first component of the optimal control $\alpha^*_1(0,x)$ (right) for $x=(x_1,0, \dots, 0)$ with $x_1  \in [-2.5, 2.5]$ for a 50-dimensional LQR problem with jumps. Orange dotted lines: numerical results of GPI-CBU with $\pm 1$ standard deviation given by orange shaded area. Blue lines: reference solutions $V$ and $\alpha^*$.}
\label{fig2}

\end{figure}

Figure \ref{fig2} shows results of GPI-CBU for the same LQR problem with jumps in $d = 50$ dimensions. While the high dimensionality renders GPI-PINN infeasible, GPI-CBU achieves high accuracy in the approximation of the value function as well as the optimal policy.
Additional results for up to 150-dimensional LQR problems are reported in Appendix \ref{lqr2}.

 \vspace{-2.5mm}
\begin{figure}[h!]  
	\centering
\hspace{-1.2mm}	\begin{minipage}{0.45\textwidth} 
		\includegraphics[width=\linewidth]{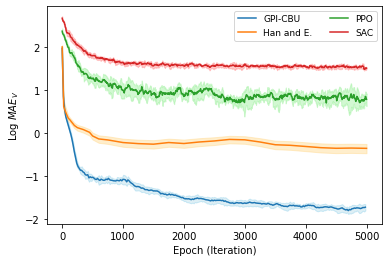} 
		\caption{\small $\log {\rm MAE}_V$ of different deep-learning methods 
		for a 10-dimensional LQR problem with jumps.} \label{fig3}
	\end{minipage}\hfill 
	\begin{minipage}{0.51\textwidth}
	\vspace{-1mm} {\small Figure \ref{fig3} shows the accuracy of GPI-CBU for 
	the 10-dimensional LQR problem with jumps from the right panel of Figure \ref{fig1}
	compared with the two popular model-free RL algorithms PPO and SAC as well as
	the model-based discrete-time approach of \cite{han2016deep} applied 
	to a time-discretization of the state dynamics \eqref{Xa}. It can be seen that in this setup,
	PPO and SAC cannot compete with the two model-based approaches since they do not 
	explicitly use the dynamics \eqref{Xa} but instead only rely on sampling from the environment. 
	The method of \cite{han2016deep} outperforms PPO and SAC but does not achieve the same 
	accuracy as GPI-CBU due to discretization errors and since it does not generalize well to unseen points in the test set. Trying to cover the space-time domain well, we ran the method of \cite{han2016deep} from several randomly sampled starting points $x_0 \in D$. 
But being a local method, it tends to learn the optimal control only along the optimal state trajectories $(t,X^{\alpha^*}_t)_{0\leq t \leq T}$, which results in poor performance in parts of the space-time domain that are not explored well. Additional results are discussed in Appendix \ref{lqr2}.}
	\end{minipage} 
	\end{figure}

\subsection{Optimal consumption-investment with jumps} 
\label{exo2}

\begin{figure}[H]
	\centering
	\small
\hspace{-0.2mm}	\begin{minipage}{.5\textwidth} \vspace{-1mm}
		\includegraphics[height=4.1cm, width=0.98\linewidth]{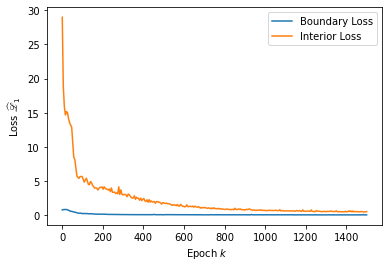}
		\label{fig:sub6}
	\end{minipage} \hspace{-2.4mm}
	\begin{minipage}{.5\textwidth}  \vspace{2mm}
		\includegraphics[height=4.1cm,width=1\linewidth]{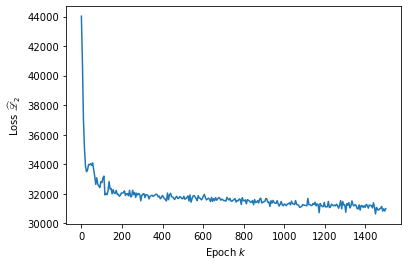}
		\label{fig:sub7}
	\end{minipage} \vspace{-5mm}
\caption{\small Training losses $\widehat{\mathscr{L}}^{(k)}_1\hspace{-0.5mm}(\theta^{(k+1)})$ (left) and  $\widehat{\mathscr{L}}^{(k+1)}_2\hspace{-0.5mm}(\phi^{(k+1)})$ (right) of GPI-CBU as functions of the epoch $k$
for a consumption-investment problem of the form \eqref{inv1}--\eqref{inv2} with $n = 25$ financial assets. 
The blue curve in the left plot represents the interior  loss of $\widehat{\mathscr{L}}^{(k)}_1$ and the orange curve its boundary part; see \eqref{How3}.} 
\label{fig:main4}
\end{figure}

As a second example, we consider an economic agent who consumes at relative rate 
$c_t$ and invests in $n$ financial assets according to a strategy\footnote{$\pi^i_t$ 
describes the fraction of the agent's total wealth held in the $i^{th}$ asset at time $t$.} $(\pi^1_t, \dots, \pi^n_t)$
so as to maximize 
\begin{equation} \label{inv1}
\E \edg{\int_0^T e^{- \rho s} u(c_s Y^\alpha_s) ds + e^{-\rho T}U(Y^\alpha_T)} 
\end{equation}
for two utility functions\footnote{In our numerical experiments, we choose CRRA utility functions.} 
$u, U \colon \R_+ \to \R$, 
 where $Y^\alpha_t$ is the wealth process evolving like \vspace{-0.3mm}
 \begin{equation} \label{inv2}
 \frac{d Y^\alpha_t}{Y^\alpha_{t-}} 
	= \brak{r_t + \sum_{i=1}^n \pi^i_t(\mu^i - r_t) - c_t} dt 
	+ \sum_{i=1}^n \pi^i_t \edg{\sigma^i_t  \, dW^i_t + d \sum_{j=1}^{M^i_t} \brak{e^{Z^i_j} - 1}} 
\end{equation} for a stochastic interest rate $r_t$, expected return rates $\mu^i$, stochastic 
volatilities $\sigma^i_t$, 
an $n$-dimensional Brownian motion $W$ with correlation matrix $\Sigma$, Cox processes $M^i$ with 
stochastic jump intensities $\lambda^i_t$ and random jump variables $Z^i_j$.
We consider strategies of the form and $c(t,\sigma_t, \lambda_t, r_t, Y^\alpha_{t-})$
and $\pi^i(t, \sigma_t, \lambda_t, r_t, Y^\alpha_{t-})$. The problem has $d = 2n + 2$ state variables, 
consisting of $\sigma^i_t, \lambda^i_t, r_t, Y^{\alpha}_t$ and $n + 1$ decision variables
$c_t, \pi^i_t$, $i = 1, \dots, n$.
In general, the corresponding HJB equation, given in \eqref{HJBinv2} in Appendix \ref{optc2}, 
does not have an analytical solution. But it can be seen in Figure \ref{fig:main4} that GPI-CBU converges numerically
for $n = 25$ financial assets. More details about this consumption-investment  problem with stochastic coefficients are provided in Appendix \ref{optc2}.

In Appendix \ref{optcv}, we consider a simplified version of the problem where 
the coefficients $\sigma^i_t, \lambda^i_t, r_t$ are constant. This case can be reduced to a one-dimensional 
ODE that can be solved with a standard Runge--Kutta scheme to obtain reference solutions. 
GPI-CBU yields numerical solutions that are virtually indistinguishable from the Runge--Kutta results.\vspace{-2.5mm}

\section{Conclusions,  limitations and future work} 

In this paper, we have introduced two iterative deep learning algorithms for solving finite-horizon stochastic control problems with jumps. Both use an actor-critic approach and train two neural networks to approximate the value function and optimal control, providing global solutions over the entire space-time domain without requiring simulation or discretization of the underlying state dynamics. Our first algorithm, GPI-PINN, works well for high-dimensional problems without jumps but becomes computationally infeasible in the presence of jumps. The second algorithm, GPI-CBU, leverages an efficient expectation-free updating rule based on Proposition \ref{Prop2} which makes it particularly well-suited for high-dimensional problems with jumps. Both algorithms are model-based. As such, they outperform 
model-free RL methods in cases where the underlying state dynamics are known. The 
accuracy and scalability of the two algorithms has been demonstrated in different numerical examples.

A limitation of our approach lies in the need to know the underlying dynamics of the state process, which are not always available in real-world applications. While it is reasonable to assume that physical systems obey known laws of motion, dynamics in economics and finance typically need to be inferred from data. However, in such cases, they can be learned in a preliminary step using e.g. recent model-learning algorithms such as \cite{brunton2016discovering} or \cite{champion2019data}. Once an approximate model has been learned from data, one of the proposed 
algorithms, GPI-PINN or GPI-CBU, can be applied to efficiently solve the resulting stochastic control problem.  

\section*{Acknowledgments} 
Financial support from Swiss National Science Foundation Grant 10003723 is gratefully acknowledged. We thank the reviewers for their helpful comments and constructive suggestions.

\bibliographystyle{icml2022}
\bibliography{example_paper2}

\newpage
\appendix
\section{Proofs}
\subsection{Proof of Proposition \ref{Prop1}} 
\label{Proof1}
Denoting $v_i(\cdot) = v\Big( t+\frac{h^2}{2n}, x + \frac{h}{\sqrt{2}}\sigma_i(t,x,a) + \frac{h^2}{2n} \beta(t,x,a)\Big)$, the second-order derivative of $\psi(h) := \sum_{i=1}^{n} v_i (\cdot)$ is given by 
{\small	\begin{align}
\psi''(h) =& \sum_{i=1}^n \Bigg(\frac{1}{n}\partial_t v_i(\cdot) + \frac{h^2}{n^2}  \partial^2_{tt}v_i(\cdot) 
+ 2 \frac{h}{n} \left(\frac{\sigma_i(t,x,a)}{\sqrt{2}} + \frac{h}{n}\beta(t,x,a) \right)^{\top} \nabla_{tx}v_i(\cdot) \nonumber
\\ & \quad + \frac{1}{n} \beta^{\top}(t, x, a) \nabla_xv_i(t, x) + \left(\frac{\sigma_i(t,x,a)}{\sqrt{2}} + \frac{h}{n}\beta(t,x,a) \right)^{\hspace{-0.7mm}\top} \hspace{-1mm} \nabla^2_x v_i(\cdot)\left(\frac{\sigma_i(t,x,a)}{\sqrt{2}} + \frac{h}{n}\beta(t,x,a) \right) \Bigg) .\nonumber
	\end{align}}
	Evaluating it at $h=0$ gives
	\begin{equation}
	\psi''(0) =  \partial_t v(t,x) + \beta^{\top}(t, x, a) \nabla_xv(t, x) + \frac{1}{2} \normalfont \text{Tr}  \left[ \sigma \sigma^\top\hspace{-0.5mm}(t,x,a)\nabla^2_xv(t,x) \right] , \nonumber
	\end{equation}
which proves the proposition. \qed

\subsection{Proof of Proposition \ref{Prop2}} 
\label{Proof2}
Since $Y_t$ is independent of $Z_1$, one obtains from the definition of $G_{\zeta}$ that
	\begin{align}
	\mathbb{E}[G_{\zeta}(t,Y_t,  Z_1, V^\alpha, \alpha(t,Y_t)) \mid Y_t= x] = V^\alpha(t,x)   &+ \zeta \hspace{0.1mm}\Big( \partial_t V^\alpha(t,x) + f(t,x,\alpha(t,x))   \hspace{-0.5mm} \nonumber
		\\
		& \hspace{-42mm} +	\beta^\top \hspace{-0.5mm}(t, x, \alpha(t,x)) \, \nabla_x V^\alpha(t, x)  
		+ \hspace{-0.5mm}\frac{1}{2} \normalfont \text{Tr} \left[ \sigma \sigma^\top\hspace{-0.5mm}(t,x,\alpha(t,x))\nabla^2_x V^\alpha(t,x) \right] \nonumber
		\\
		& \hspace{-42mm}  + \lambda(t,x,\alpha(t,x)) \, \mathbb{E} \left[V^\alpha(t,x+\gamma(t,x,Z_1,\alpha(t,x))) - V^\alpha(t,x) \right]  \Big) \nonumber
		\\[0.1cm] \label{Bell}
		&\hspace{-17mm}= V^\alpha(t,x) + \zeta \, \mathcal{H}(t,x,V^\alpha, \alpha(t,x))\\[0.1cm]
		&\hspace{-17mm}= V^\alpha(t,x), \notag
	\end{align} 
where the last equality follows from Theorem \ref{thm:FK}. On the other hand, it is well known that 
\[
\mathbb{E}[G_{\zeta}(t,Y_t,  Z_1, V^\alpha, \alpha(t,Y_t)) \mid Y_t] = g(Y_t)
\]
for the Borel measurable function $g \colon  D \to \R$ minimizing the mean squared error
\[
\mathbb{E} \edg{ \big( g(Y_t) - G_{\zeta}(t,Y_t, Z_1, V^{\alpha}, \alpha(t, Y_t)) \big)^2 };
\]
see e.g. Theorem 4.1.15 of \citet{durrett}. This completes the proof. \qed

\section{DGM Architecture} 
\label{dgm}
\begin{figure}[H]
	\centering  
	\includegraphics[scale=0.153]{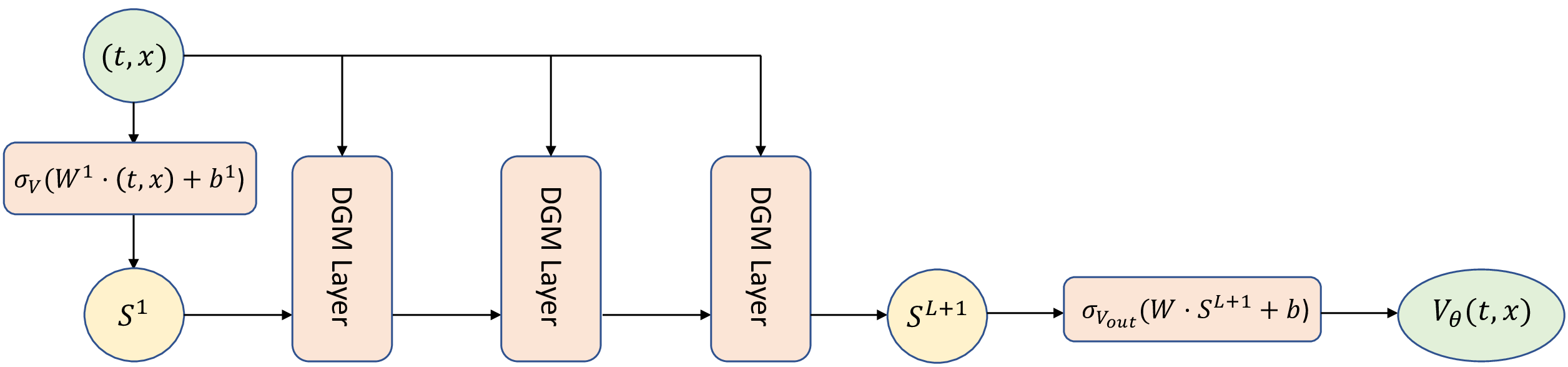} 
\caption{\small DGM architecture of the value neural network with $L=3$ (\textit{i.e.}\ 4 hidden layers).}
	\label{U_S process} 
\end{figure}
Each DGM layer in Figure \ref{U_S process} in the value network $V_\theta$ is of the following form
$$
\begin{aligned}
	S^1 & =\sigma_V\left(W^1 \cdot (t,x)+b^1\right), \\
	Z^{\ell} & =\sigma_V\left(U^{z, \ell} \cdot (t,x)+W^{z, \ell} \cdot S^{\ell}+b^{z, \ell}\right), \quad \ell=1, \ldots, L, \\
	G^{\ell} & =\sigma_V\left(U^{g, \ell} \cdot (t,x)+W^{g, \ell} \cdot S^1+b^{g, \ell}\right), \quad \ell=1, \ldots, L, \\
	R^{\ell} & =\sigma_V\left(U^{r, \ell} \cdot (t,x)+W^{r, \ell} \cdot S^{\ell}+b^{r, \ell}\right), \quad \ell=1, \ldots, L, \\
	H^{\ell} & =\sigma_V\left(U^{h, \ell} \cdot (t,x)+W^{h, \ell} \cdot \left(S^{\ell} \odot R^{\ell}\right)+b^{h, \ell}\right), \quad \ell=1, \ldots, L, \\
	S^{\ell+1} & =\left(1-G^{\ell}\right) \odot H^{\ell}+Z^{\ell} \odot S^{\ell}, \quad \ell=1, \ldots, L, \\
	V_\theta(t, x) & =\sigma_{{V}_{out}} \left(W \cdot S^{L+1}+b \right), 
\end{aligned}
$$
where the number of hidden layers is $L+1$, $\cdot$ denotes matrix multiplication 
and $\odot$ element-wise multiplication. The DGM parameters of the value network are
\begin{align*}
		\theta = \Big\{ \hspace{-0.5mm}W^1, b^1, \left(U^{z, \ell}, W^{z, \ell}, b^{z, \ell}\right)_{\ell=1}^L,&\left(U^{g, \ell}, W^{g, \ell}, b^{g, \ell}\right)_{\ell=1}^L,
		\\
	&\left(U^{r, \ell}, W^{r, \ell}, b^{r, \ell}\right)_{\ell=1}^L, \left(U^{h, \ell}, W^{h, \ell}, b^{h, \ell}\right)_{\ell=1}^L , W, b\Big\} .
\end{align*}The number of units in each layer is $N$.  $\sigma_V: \mathbb{R}^N \rightarrow \mathbb{R}^N$ is a twice-differentiable element-wise nonlinearity and $\sigma_{V_{out}}: \mathbb{R}^N \to \mathbb{V}$ is the output activation function, also twice-differentiable, where $\mathbb{V} \subseteq \R$ is chosen so as to satisfy possible restrictions on the value function's output, resulting from the form of the stochastic control problem (e.g.\ non-negative value function). The control network is designed following the same DGM architecture with $\sigma_{\alpha_{out}} : \mathbb{R}^N \rightarrow A$. Throughout the numerical examples of the paper, we use $L=3$ with $N = 50$ neurons in 
each of the DGM layers, see Figure \ref{U_S process}. In the value network, we use $\tanh$ for the 
activation function $\sigma_V$ and softplus for $\sigma_{V_{out}}$. For the control network, we adopt $\tanh$ for 
$\sigma_\alpha$, while as output activation $\sigma_{\alpha_{\text{out}}}$ we choose the identity in Example~\ref{Sec:LQR} and softplus in Example~\ref{exo2}.

Unless stated otherwise, we use a number of sample points $M_1, M_2$  equal  to $256$, a number of gradient steps $N_1, N_2$ equal to $64$ and a maximum number of epochs $k_{*} = 1500$. The network parameters are updated using Adam \citep{adam} with constant learning rates $\eta_1, \eta_2$ equal to $0.001$. Both GPI-PINN and GPI-CBU
are fairly robust with respect to these choices. In our setting, the most critical hyperparameters for convergence are the proportionality factors $\xi_1, \xi_2$, which we determined using the approach of \cite{wang2022and}, together with the scaling rate $\zeta$ (specific to GPI-CBU). In our experiments, we set $\zeta =1$ as it provides a good trade-off between convergence speed and accuracy of the improvements in GPI-CBU. Alternatively, one could consider an adaptive scaling factor $\zeta_k > 0$ depending on the epoch $k$. On the other hand, 
negative scaling factors resulted in non-convergence in all our experiments with 
exploding losses $\widehat{\mathscr{L}}_1^{(k)}$ and $\widehat{\mathscr{L}}_2^{(k)}$ after only a few epochs. 

Algorithms \ref{Algo1} and \ref{Algo2} were implemented using TensorFlow and Keras
with GPU acceleration on an NVIDIA RTX 4090.
\section{Linear-quadratic regulator with jumps: detailed results} 
\label{lqr2}
The value function of the LQR problem with jumps \eqref{LQR} is given by 
\begin{equation}
V(t,x) = 	\inf_{\alpha} \mathbb{E}\left[ \int_0^T c_1\, \|\alpha_s\|^2_2 \, ds + c_2 \|X^\alpha_T \|^2_2  \  \Big| \, X^\alpha_t =x \right] . \label{vv2} \vspace{-0.1mm}
\end{equation} The corresponding  HJB equation is 
\begin{equation*} \label{HJB3} 
\begin{aligned}
	&0= \partial_t V(t,x) + \frac{1}{2}\text{Tr} \left[B B^\top \hspace{-0.3mm}\nabla^2V_{x}(t,x)\right] \hspace{-0.4mm} +\inf_{a \in \mathbb{R}^d} \Big\{ c_1 \|a\|^2_2 + a^\top \nabla_x V(t,x)
	\\[-0.6mm]
	& \hspace{30mm}+ ( \Lambda_1+ \Lambda_2 \|a\|^2_2) \, \mathbb{E} \left[V(t, x +  Z_1)  -  V(t,x) \right] \Big\}  
\end{aligned}
\end{equation*}
with terminal condition $V(T,x) = c_2 \|x\|^2_2$. Using the Ansatz 
\[
V(t,x) = g(t) + \frac{1}{2} h(t) \|x\|^2_2 \, ,
\]
it is straightforward to see that $g$ is of the form
\begin{equation} \label{solg}
g(t) = \frac{1}{2} \big( \text{Tr}[B B^\top]  +\Lambda_1  \hspace{0.1mm} \upsilon) \int_t^T h(s) ds 
\end{equation}
for the unique solution of the non-linear ODE
\begin{equation} \label{ODEh}
h'(t) = \frac{h^2(t)}{2c_1 + h(t) \hspace{0.2mm} \upsilon \hspace{0.2mm}\Lambda_2}, \quad h(T) = 2 c_2\, .
\end{equation}
\eqref{ODEh} can efficiently be solved using a numerical method such as Runge--Kutta of order $5(4)$; see e.g.
\cite{dormand1980family}. The optimal control $\alpha^*$ is given in terms of $h$ by
\begin{equation*} \label{analy2}
\alpha^*(t,x) = - \frac{h(t)x}{2 c_1 + h(t) \hspace{0.2mm} \upsilon \hspace{0.1mm}\Lambda_2} \, .
\end{equation*}
Note that in the constant intensity case $\Lambda_1 > 0, \Lambda_2 = 0$, 
\eqref{solg}--\eqref{ODEh} admit the explicit solutions
\begin{equation} \label{gh}
g(t) = (\text{Tr} [B B^T ] + \Lambda_1 v)\,c_1\left( \log(\frac{c_2}{c_1}(T-t)+1)\right) 
\quad \mbox{and} \quad h(t) = \frac{2 c_1 c_2}{c_1+ c_2 (T-t)} \,.
\end{equation}
But also in the case $\Lambda_2 > 0$, numerical solutions of \eqref{solg}--\eqref{ODEh} 
obtained with Runge--Kutta of order 5(4) provide very precise reference solutions 
$V$ and $\alpha^*$ to the LQR problem \eqref{LQR}.

GPI-PINN and GPI-CBU both need random points $(t_m, x_m) \in [0,T) \times \R^d$ and 
$(y_m) \in \R^d$ sampled from two distributions $\mu$ on $[0,T) \times \R^d$ and $\nu$
on $\R^d$, which we chose according to \cite{bachouch2022deep}. 

\begin{figure}[H]
	\centering
	\subfigure{
		\includegraphics[width=0.48\columnwidth]{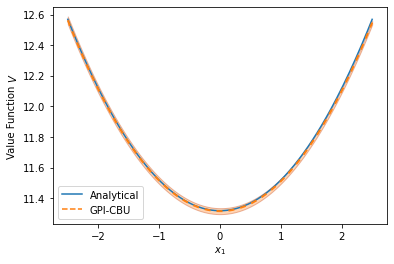}
		\label{fig:sub1}
	} \hspace{-3mm} 
	\subfigure{ 
		\includegraphics[width=0.483\columnwidth ]{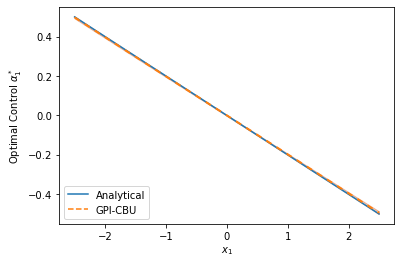}
		\label{fig:sub2}
	}  \vspace{-2mm}
\caption{\small Value function $V(0,x)$ (left) and first component of the optimal control $\alpha^*_1(0,x)$ (right) for $x=(x_1,0, \dots, 0)$ with $x_1 \in [-2.5, 2.5]$. Orange dotted lines: numerical results of GPI-CBU with $\pm 1$ standard deviation given by orange shaded area. Blue lines: analytical solution \eqref{gh}.
$d = 10$, $\Lambda = 1/4$, $\Lambda_2 = 0$, $k_{*}= 1500$.} \vspace{-4mm}
	\label{fig:main}
\end{figure}

\begin{figure}[H]
	\centering
	\begin{minipage}{.5\textwidth}
		\includegraphics[width=0.95\linewidth]{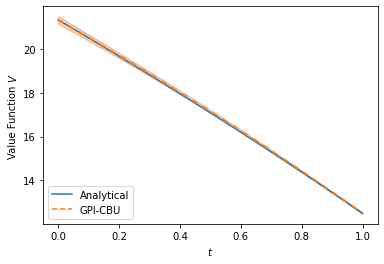}
		\label{fig:sub3}
	\end{minipage} \hspace{-3mm}
	\begin{minipage}{.5\textwidth} \vspace{4.5mm}
		\includegraphics[width=1\linewidth]{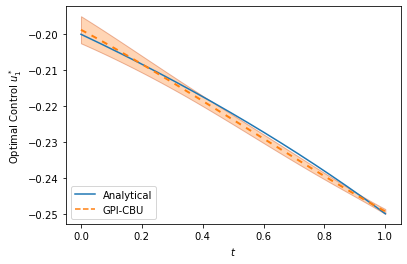}
		\label{fig:sub4}
	\end{minipage}  \vspace*{-7mm}
	\caption{\small Value function $V(t,x)$ (left) and first component of the optimal control $\alpha^*_1(t,x)$ (right) for  $t \in [0,1]$ and $x= \mathbf{1}_{50}$. Orange dotted lines: numerical results from GPI-CBU with $\pm 1$ standard deviation in orange shaded area. Blue lines: analytical solution \eqref{gh}. $d=50$, 
$\Lambda_1 = 1/4$, $\Lambda_2=0$, $k_{*}= 1500$.}
\label{fig:main2}
\end{figure}

\begin{figure}[H]
	\includegraphics[scale=0.534]{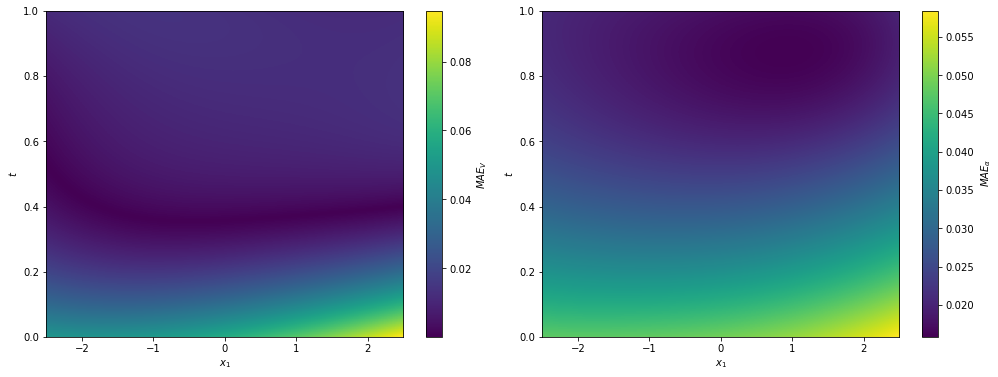}
	\vspace{-2mm}
	\label{fig:sub5}  \vspace*{-3mm}
\caption{\small Heatmaps of ${\rm MAE}_V$ (left) and ${\rm MAE}_\alpha$ (right) for $t\in [0,1]$ and 
$x=(x_1, 1, \dots, 1)$ with $x_1  \in [-2.5, 2.5]$ obtained from GPI-CBU for the LQR problem 
with $d=50$, $\Lambda_1 = 1/4$ and $\Lambda_2=0$. $k_* = 1500$.}  \vspace{-3mm}
	\label{heatmap1}
\end{figure}

In the following we consider problem \eqref{LQR} with $T = 1$, $B= I_d$, $c_1 = 1$, $c_2 = 1/4$ and
$d$-dimensional standard normal jumps $Z_j$. Figures \ref{fig:main}--\ref{heatmap1} show results 
obtained with GPI-CBU in $d=50$ dimensions for the constant jump intensity case 
$\Lambda_1 = 1/4$ and $\Lambda_2 = 0$, which admits analytical reference solutions.

  \vspace{-2mm}
 \vspace{-2mm}
 \vspace{-2mm}

${\rm MAE}_\alpha$ in Figure \ref{heatmap1} is defined, analogously to ${\rm MAE}_V$
in Section \ref{Sec:LQR}, by \vspace*{-1mm}
\[
{\rm MAE}_\alpha = \frac{1}{M} \sum_{i=1}^M \| \alpha_{\phi^{(k_{*})}}\hspace{-0.4mm}(t_i, x_i) - \alpha^*(t_i, x_i) \|_2
\vspace*{-1mm}
\]
for a test set of size $M$ uniformly sampled from $[0,1] \times [-2.5,2.5]^d$. 

\begin{figure}[h!]
	\centering
	\begin{minipage}{.5\textwidth}
		\includegraphics[width=0.94\linewidth]{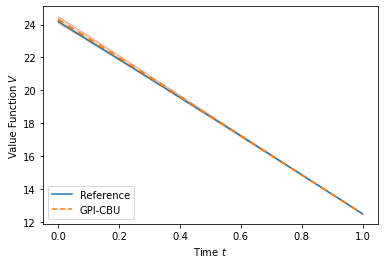}
		\label{fig:test101} \vspace{6mm}
	\end{minipage} \hspace{-4.5mm}
	\begin{minipage}{.5\textwidth}
	\vspace{-1.7mm}	\includegraphics[width=1.01\linewidth]{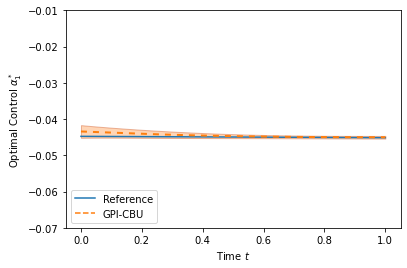}
		\label{fig:test210}
	\end{minipage}\vspace*{-6mm}
	\caption{\small Value function $V(t,x)$ (left) and first component of the optimal control $\alpha^*_1(t,x)$ (right)
	for $t \in [0,1]$ and $x= \mathbf{1}_{50}$. Orange dotted lines: numerical results of GPI-CBU with $\pm 1$ standard deviation in orange shaded area. Blue lines: reference solution.
$d= 50$, $\Lambda_1 = 0$, $\Lambda_2 = 2$, $k_* = 1500$.}
	\label{comp411}
\end{figure}

\begin{figure}[h!]
	\centering
	\begin{minipage}{.5\textwidth}
		\includegraphics[width=0.96\linewidth]{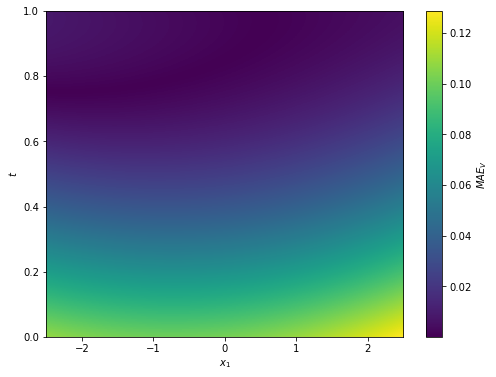}
		\label{fig:testt1}
	\end{minipage}%
	\begin{minipage}{.5\textwidth}
		\includegraphics[width=0.96\linewidth]{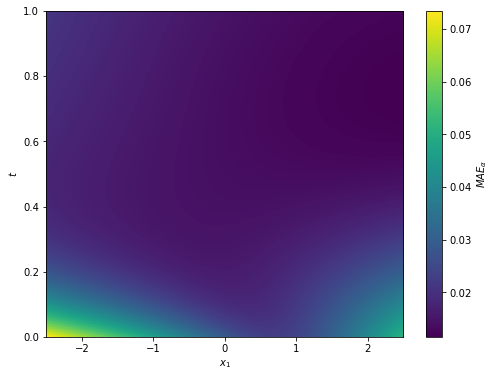}
		\label{fig:testt2}
	\end{minipage}  \vspace{-1mm}
\caption{\small Heatmaps of ${\rm MAE}_V$ (left) and ${\rm MAE}_\alpha$ (right) for $t\in [0,1]$ and $x=(x_1,\bm{1}_{49})$ with $x_1  \in [-2.5, 2.5]$ obtained from GPI-CBU for the LQR problem 
with $d=50$, $\Lambda_1 = 1/4$ and $\Lambda_2=0$. $k_* = 1500$.} 
	\label{heatmap5}
\end{figure}

Next, we consider the same LQR problem \eqref{LQR} as above 
in 50 dimensions with a controlled jump intensity. 
Figures \ref{comp411}--\ref{heatmap5} (and \ref{fig2}), show results of GPI-CBU for 
$\Lambda_1 = 0$ and $\Lambda_2 = 2$.
Reference solutions were computed with Runge--Kutta of order 5(4).

\vspace{-3.5mm}
 \newpage
These results illustrate the accuracy of GPI-CBU for high-dimensional control problems with 
controlled jumps. It can be seen that the optimal control $\alpha^*$ is lower in magnitude 
compared to the case of constant jump intensity since now, exerting control increases the 
likelihood of jumps. But in both cases, it is optimal to steer the state process 
$X^\alpha_t$ towards $0$ as $t$ approaches the time horizon $T$.

Table \ref{tab:highdim_results} summarizes results of GPI-CBU for higher-dimensional LQR problems 
with the same parameters as above with controlled jump intensity $\Lambda_1 =0$, 
$\Lambda_2 = 2$ obtained over $k_* = 5000$ training epochs. 
This demonstrates the scalability of GPI-CBU. 
\begin{table}[h!]
\caption{\small GPI-CBU performance metrics as a function of the state dimension $d$ for LQR problems 
of the form \eqref{LQR} with $\Lambda_1 = 0$, $\Lambda_2 = 2$ and $k_* = 5000$ training epochs.} \vspace*{2mm}
	\label{tab:highdim_results}
	\centering
	\begin{tabular}{c|c|c|c|c|c}
		Dimensions $d$ & ${\rm MAE}_V$ & ${\rm MAE}_\alpha$ & Loss $\widehat{\mathscr{L}}_1(\theta^{k_*}) $ & Loss $\widehat{\mathscr{L}}_2(\theta^{k_*}) $ & Time (sec) \\[1mm]
		\hline \\[-3mm]
		5   & 0.0023 & 0.0041  & 0.0237 & -0.1332 & 6,410 \\
		10  & 0.0025 & 0.0049  & 0.0314 & -0.1972 & 8,093 \\
		50  & 0.0147 & 0.0075  & 0.1267 & -0.4206 & 16,129 \\
		100 & 0.0492 & 0.00539 & 0.6970 & -0.4461 & 24,359 \\
		150 & 0.0979 & 0.0096  & 3.7210 & -0.4671 & 33,120 \\
		\hline 
	\end{tabular}
\end{table}

\section{Optimal consumption-investment problem}
\subsection{Constant coefficients} 
\label{optcv}
We first study the optimal consumption-investment problem in a model 
with constant coefficients, where the risk-free asset evolves according to
\begin{equation}
	d S^0_t = r S^0_t dt  \, , \label{r}
\end{equation}
for a constant interest rate $r \in \R$, and there are $n$ stocks with prices following
\begin{equation}
	\frac{dS^i_t}{S^i_{t-}} = \mu^i dt + \sigma^i dW^i_t 
	+ d \sum_{j=1}^{M^i_t} \brak{e^{Z^i_j} - 1} , \label{S}
\end{equation}
for constant drifts $\mu^i \in \R$ and volatilities $\sigma^i \in \R_+$,
an $n$-dimensional Brownian motion $W$ with correlation matrix $\Sigma$, independent Poisson processes $M^i$ with constant
intensities $\lambda^i \ge 0$ and i.i.d.\ normal random variables $Z^i_1, Z^i_2, ... \sim \mathcal{N}(\mu^i_Z, \sigma^i_Z)$ with $\mu^i_Z \in \mathbb{R}$ and $\sigma^i_Z \in \mathbb{R}_+$. Suppose an investor starts 
with an initial endowment of $Y_0 > 0$, consumes at rate $c_t Y^\alpha_{t-}$ and invests 
$\pi^i_t Y^\alpha_{t-}$, $i =1, \dots n$, in the $n$ stocks. If the risk-free asset is used to 
balance the transactions, the resulting wealth evolves according to 
\begin{equation} \label{dynconst}
	\frac{d Y^\alpha_t}{Y^\alpha_{t-}} 
	= \brak{r + \sum_{i=1}^n \pi^i_t(\mu^i - r) - c_t} dt 
	+ \sum_{i=1}^n \pi^i_t \edg{\sigma^idW^i_t + d \sum_{j=1}^{M^i_t} \brak{e^{Z^i_j} - 1}}.
\end{equation}
Let us assume the investor attempts to maximize
\[
\E \edg{\int_0^T e^{- \rho s} u(c_sY^\alpha_s) ds + e^{-\rho T} U(Y^\alpha_T)} 
\]
for two utility functions $u,U \colon \R_+ \to \R$ and a discount factor 
$\rho \in \mathbb{R}$. Since the driving noise 
in \eqref{dynconst} has stationary and independent increments,
it is enough to consider strategies of the form 
$c_t = c(t, Y^\alpha_{t-})$ and $\pi^i_t = \pi^i(t, Y^\alpha_{t-})$ for functions $c, \pi^i \colon [0,T) \times \mathbb{R}_+ \to \R_+$, $i = 1, \dots, n$. We write this $n+1$-dimensional control as $\alpha_t = (c_t, \pi^1_t, \ldots, \pi^n_t)^\top  \hspace{-1mm}\in  A := \R_+^{n+1}$. 
Assuming constant relative risk aversion (CRRA) utility functions $u, U$ with relative risk aversion $\gamma >0$ and denoting $\delta = 1-\gamma$, 
the resulting reward functional is 
\begin{equation}	
V^\alpha(t,y) = \mathbb{E}\left[\int_{t}^{T}e^{-\rho(s-t)}\frac{\left(c_{s}Y^\alpha_{s}\right)^{\delta}}{\delta}ds+e^{-\rho(T-t)}\frac{\left(Y^\alpha_{T}\right)^{\delta}}{\delta} \, \Big| \,Y^\alpha_{t}= y \right] .  \label{optinvesst}
\end{equation}
Finally, writing $\mu = (\mu^1, \ldots, \mu^n)^\top$\hspace{-0.5mm}, $\sigma = \text{diag}(\sigma^1,\ldots, \sigma^n)$, $\mu_Z = (\mu_Z^1, \ldots, \mu_Z^n)^\top$\hspace{-0.5mm}, $\sigma_Z = (\sigma^1_Z, \ldots, \sigma^n_Z)^\top$\hspace{-0.5mm}, $\lambda = (\lambda^1, \ldots, \lambda^n)^\top$, $\pi = (\pi^1, \ldots, \pi^n)^\top\hspace{-0.5mm}$, the associated HJB equation for the value function $V(t,y) = \sup_{\alpha} V^{\alpha}(t,y)$ satisfies\footnote{Note that the term $- \rho V(t,y)$ 
in \eqref{HJBinv} is not contained in our general HJB equation \eqref{HJB}. But GPI-PINN and 
GPI-CBU still work if it is added.}  for all $(t,y) \in [0,T) \times \mathbb{R}_+$,
\begin{equation} \label{HJBinv}
\begin{aligned}
	0 = \hspace{0.4mm} & \partial_t V(t,y) - \rho V(t,y) + \sup_{(c, \pi) \in A} \bigg\{ (r+(\mu-r)^{\top} \pi -c) \, y \, \partial_y V(t,y)  
	\\
	&\hspace{-2.2mm} +\frac{1}{2}  \pi^\top \hspace{-0.5mm}\sigma \Sigma \sigma^\top \hspace{-0.5mm} \pi \, y^2 \, \partial^2_{yy}V(t,y) + \sum_{i=1}^{n} \lambda^i \mathbb{E}\left[V(t, y+y \, \pi^{i} (e^{Z^i_1} - 1)) - V(t,y) \right] +\frac{(cy)^\delta}{\delta}\bigg\} \, , 
\end{aligned}
\end{equation}

with $V(T,y) = y^\delta/ \delta$. This stochastic control problem does not admit an analytical solution but we can instead characterize its solution in terms of a PIDE, so as to assess the accuracy of the proposed Algorithms \ref{Algo1} and \ref{Algo2}. Following \citet{oksendal2007applied}, we assume the value function is of the form 
$V(t,y)  = A(t) y^\delta / \delta$. Then optimizing the HJB equation \eqref{HJBinv} leads to first order conditions
showing that the optimal consumption rate $c^*$ is independent from the wealth $y$ and given by
\begin{equation}
c^*(t,y) = A(t)^{1/(\delta -1)} \label{Ceq},
\end{equation}
and the optimal investment strategy is given by a 
vector of constants $\pi^*(t,y) = \pi^* :=(\pi^{*,1}, \ldots,\pi^{*,n})^\top $ satisfying
\begin{equation*}
	(\mu-r)^\top + \sigma \Sigma \sigma^\top \pi^*  (\delta -1) + \sum_{i=1}^{n} \lambda^i \, \mathbb{E} \left[ \left(1+\pi^{*, i} (e^{Z^i_1} - 1)\right)^{\delta-1} \left(e^{Z^i_1} - 1\right)  \right] = 0.
\end{equation*}
Plugging these optimal controls back into the HJB equation \eqref{HJBinv}, 
we find that the function $A(t)$ satisfies the ODE
\begin{equation} \label{Aeq}
\begin{aligned}
	\frac{	\partial_t A(t)}{A(t)} = \rho \hspace{0.5mm}- & \delta \left(r+(\mu-r)^{\top} \pi^* - A(t)^{1/(\delta-1)}\right) - \frac{1}{2} \delta (\delta-1) \pi^{*\top}\sigma \Sigma \sigma^\top \pi^* 
	\\[-1mm]
	&\hspace{21mm}- \sum_{i=1}^{n} \lambda^i \mathbb{E} \left[ \left(1+\pi^{*,i} (e^{Z^i_1} - 1)\right)^{\delta} \hspace{-0.5mm}-1 \right] - A(t)^{1/(\delta-1)} ,
\end{aligned}
\end{equation}
with $A(T) = 1$, which can be solved numerically using again the Runge--Kutta method of order 5(4). For both Algorithms \ref{Algo1} and \ref{Algo2}, we sample the time and space points independently from the uniform distributions $\mathcal{U}_{[0,T]}$ and $\mathcal{U}_{[0,y_b]}$, respectively.  The parameters of the optimal investment problem below are as follows: $T=1, y_b=150, r=0.02, \rho= 0.045, \delta =0.7, \lambda  = 0.45\cdot  \mathbf{1}_n, \mu_Z  = 0.25  \cdot\mathbf{1}_n, \sigma_Z  = 0.2 \cdot \mathbf{1}_n, \mu = 0.032  \cdot\mathbf{1}_n,$ $\sigma =I_n$, $\Sigma = 0.2 \cdot \mathbf{1}_{n\times n}$ with $\text{diag}(\Sigma) = \mathbf{1}_n$. Note that, compared to the LQR problem in Section \ref{Sec:LQR}, the proportionality factor $\xi$ is now set to $10$. Moreover, since $A = 
\R_+^{n+1}$, $\sigma_{\alpha_{out}}$ is chosen to be the softplus activation function. Instead of  the DGM architecture, we here train a classical feedforward neural network with 4 hidden layers, each of 128 neurons. Finally, the ${\rm MAE}_V$ and ${\rm MAE}_\alpha$ metrics are again defined as in Section \ref{Sec:LQR} on a test set of size $M$, uniformly sampled from $[0,1] \times [0, y_b]$ with $V$ and $\alpha^*$ obtained from the Runge--Kutta method described above.

\begin{figure}[H]
	\centering
	\begin{minipage}{.5\textwidth}
		\includegraphics[width=0.92\linewidth]{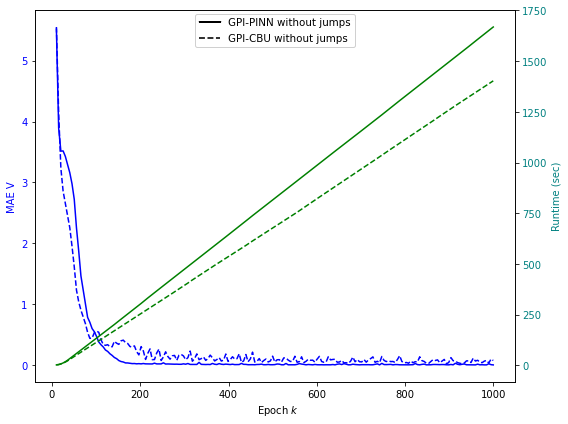}
		\label{fig:test214}
	\end{minipage}%
		\begin{minipage}{.5\textwidth}
	\vspace{4mm}
	\includegraphics[width=0.992\linewidth]{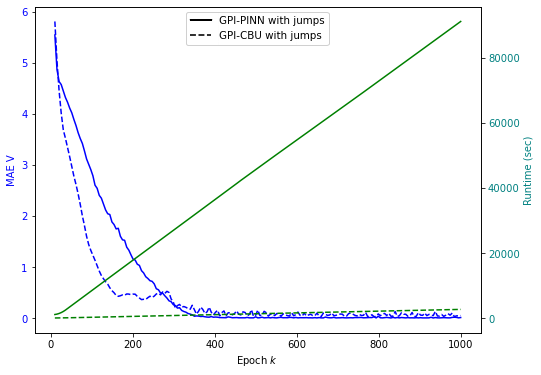}
	\label{fig:test114}
	\end{minipage}\vspace*{-5mm}
\caption{\small Comparison of ${\rm MAE}_V$ (blue) and runtime in seconds (green) of GPI-PINN  (solid line) and  GPI-CBU (dashed line) as functions of the number of epochs $k$ for the optimal consumption-investment problem without jumps (left) and with jumps (right) for $n=10$ stocks.\vspace{-4mm}} 
	\label{compk}
\end{figure}

\begin{figure}[H]
	\centering
	\begin{minipage}{.5\textwidth}
		\includegraphics[width=0.96\linewidth]{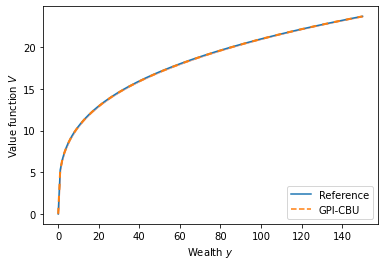}
		\label{fig:test1c}
	\end{minipage}%
	\begin{minipage}{.5\textwidth}
		\includegraphics[width=0.96\linewidth]{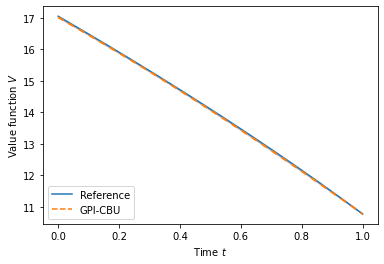}
		\label{fig:test2c}
	\end{minipage}\vspace*{-1mm}
\caption{\small Value function $V(t,y)$ for $t=0$, $y \in [0,150]$ (left) and $t\in [0,1]$, $y=50$ (right) for the optimal consumption-investment problem \eqref{inv1} with constant coefficients  and $n=50$ stocks. Orange dotted lines: results of GPI-CBU with $\pm$ 1 standard deviation given by orange shaded area ($k_{*}= 1000$). Blue lines: reference solution from  Runge--Kutta applied to \eqref{Ceq}-\eqref{Aeq}.}
	\label{comp4}
\end{figure}

\begin{figure}[H]
	\begin{center}
		\centerline{\includegraphics[scale=0.54]{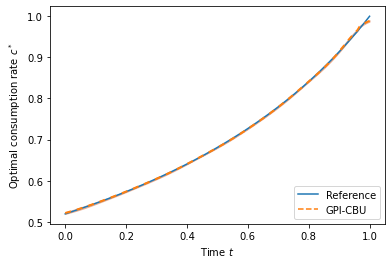}} \vspace{-1mm}
\caption{\small Optimal relative consumption rate  $c^*(t)$ for $t \in [0,1]$ for the optimal consumption-investment problem \eqref{inv1} with constant coefficients  and $n=50$ stocks. Orange dotted line:  results of GPI-CBU with $\pm$ 1 standard deviation given by orange shaded area ($k_{*}= 1000$). Blue line: reference solution from  Runge--Kutta  applied to \eqref{Ceq}-\eqref{Aeq}.}
		\label{comp5}
	\end{center}
	\vskip -0.2in
\end{figure}

\begin{figure}[H]
	\centering
	\begin{minipage}{.5\textwidth}
		\includegraphics[width=0.96\linewidth]{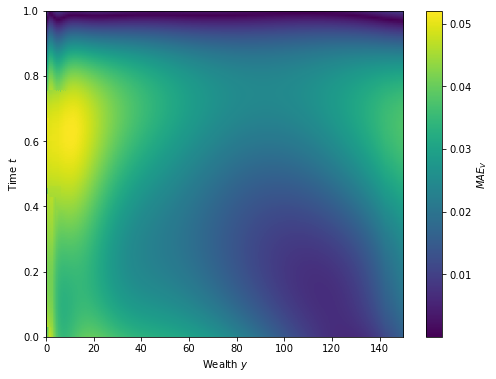}
		\label{fig:test1}
	\end{minipage}%
	\begin{minipage}{.5\textwidth}
		\includegraphics[width=0.96\linewidth]{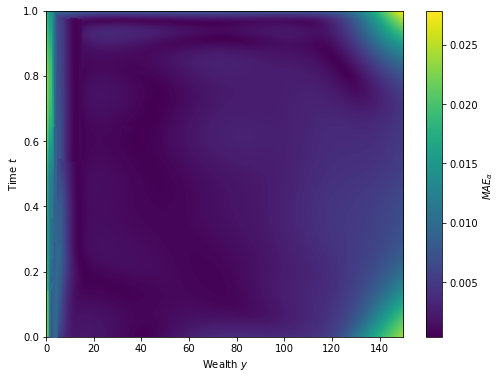}
		\label{fig:test2}
	\end{minipage}  \vspace*{-1mm}
\caption{\small Heatmaps of ${\rm MAE}_V$ (left) and ${\rm MAE}_\alpha$ (right) for $t\in [0,1]$ and  $y  \in [0, 150]$, for GPI-CBU applied to the optimal consumption-investment problem \eqref{inv1} with constant coefficients and $n=50$ stocks.}  
	\label{heatmap2}
\end{figure}

In Figure \ref{compk} we again compare ${\rm MAE}_V$ and runtime of GPI-PINN and GPI-CBU as functions of the number of epochs $k$ for  $n=10$ stocks in the consumption-investment problem \eqref{optinvesst} with and without jumps. We again see that the residual-based GPI-PINN Algorithm \ref{Algo1} tends to be more stable and accurate for larger epochs, despite having a higher runtime. When accounting for jumps in the dynamics  \eqref{dynconst}, GPI-PINN   becomes numerically very time-consuming, even for a 10-dimensional control problem. This issue is  even amplified compared with the LQR problem (see Figure \ref{fig1}), as the jump size now depends on the control $\pi$ in the wealth dynamics \eqref{dynconst}. In contrast,  GPI-CBU again handles these jumps far more efficiently. \vspace{-4mm}
Next, we address a higher-dimensional optimal consumption-investment problem with jumps ($n=50$ stocks), where only GPI-CBU is implemented since GPI-PINN becomes then numerically infeasible. Figures \ref{comp4} and \ref{comp5} again confirm the accuracy of GPI-CBU for both the value function and the optimal consumption rate $c^*(t)$,  with the standard deviation across 10 independent runs of GPI-CBU being virtually imperceptible.
The optimal wealth allocations $\pi^*$, being constant, are not depicted. However, the ${\rm MAE}_\alpha$ heatmap in Figure \ref{heatmap2} corroborates our method's accuracy in determining both  $c^*$ and $\pi^*$. We also observe that the higher absolute errors tend to occur at the boundaries of the domain.  \vspace{-0.1cm}
\vspace{-3mm}

\subsection{Stochastic coefficients} 
\label{optc2}
Now, we study a more complex optimal consumption-investment problem in a realistic market of the form
\eqref{r}--\eqref{S} with a stochastic interest rate
\[
dr_t = a_r(b_t - r_t) dt + \nu_r \sqrt{r_t} \,  dW^r_{t},
\]
with stochastic (Heston) volatility models for the $n$ stock prices
\begin{align*}
	\frac{dS^i_t}{S^i_{t-}} &= \mu^i dt + \sigma^i_t \,  dW^{S,i}_t 
	+ d \sum_{j=1}^{M^i_t} \brak{e^{Z^i_j} - 1} \, ,\label{S2}	
\end{align*}
for $\sigma^i_t = \sqrt{v^i_t}$, where $v^i_t$ follows 
\begin{align}
	d v^i_t &= a^i_{v} (b^i_{v} - v^i_t) + \nu^i_{v} \sqrt{v^i_t}   \, 
	dW^{v,i}_{t},
\end{align}
and with stochastic jump intensities
\[
d \lambda^i_t = a^i_{\lambda} (b^i_{\lambda} - \lambda^i_t) + \nu^i_{\lambda} \sqrt{\lambda^i_t}  \,  
dW^{\lambda,i}_{ t} \, .
\]
We denote $W = (W^{S,1}, \ldots,W^{S,n}, W^{v,1}, \ldots, W^{v,n}, W^{\lambda,1}, \ldots, W^{\lambda,n}, W^{r})^\top$ a  $(3n+1)$-dimensional Brownian motion with correlation matrix $\Sigma$. In this case, the wealth process can still be described as follows
\begin{equation*} \label{dynconst2}
	\frac{d Y^\alpha_t}{Y^\alpha_{t-}} 
	= \brak{r_t + \sum_{i=1}^n \pi^i_t(\mu^i - r_t) - c_t} dt 
	+ \sum_{i=1}^n \pi^i_t \edg{ \sqrt{v^i_t} \  dW^{S,i}_t + d \sum_{j=1}^{M^i_t} \brak{e^{Z^i_j} - 1}} ,
\end{equation*}
and the strategies should be of the form $c(t,v_t, \lambda_t, r_t, Y^\alpha_{t-})$ and 
$\pi^i(t,v_t, \lambda_t, r_t, Y^\alpha_{t-})$, with $\alpha_t = (c_t, \pi^1_t, \ldots, \pi^n_t)^\top  \hspace{-1mm}\in  A := \R_+^{n+1}$. Hence, the dynamics of the $(2n +2)$-dimensional process $X^\alpha_t = (v_t, \lambda_t, r_t, Y^\alpha_{t-})$  are given by \vspace{-1mm}
\begin{equation*}
	dX^\alpha_t = \mu_X(t, X^\alpha_t) dt + \Sigma_X(t,X^\alpha_t) dW_t + \gamma_X(t,X^\alpha_t)   \, d \sum_{i=1}^{n} \sum_{j=1}^{M^i_t} \brak{e^{Z^i_j} - 1}  ,
\end{equation*}
 with $\mu_X(\cdot) \in \mathbb{R}^{2n +2}, \Sigma_X(\cdot) \in \mathbb{R}^{(2n +2)\times(3n+1)}_+$ and $\gamma_X(\cdot) \in \mathbb{R}^{2n +2}$ such that 
{\small \begin{equation*}
		\mu_X(t, X^\alpha_t)=\left(\begin{array}{c}
			a_v(b_v- v_t) \\
			a_\lambda(b_\lambda - \lambda_t)\\
			a_r(b_r - r_t) \\
			Y^\alpha_{t-}(r_t +(\mu - r_t)^\top \pi_t - c_t)
		\end{array}\right), \, \ \	\gamma_X(t, X^\alpha_t)=\left(\begin{array}{c}
		\bm{0}_n \\
		\bm{0}_n \\
		0 \\
		Y^\alpha_{t-} \pi_t^\top \mathbf{1}_n  \
		\end{array}\right) ,
\end{equation*}} 
and \begin{equation*}
\Sigma_X(t, X^\alpha_t) =
\begin{pmatrix}
	\mathbf{0}_{n \times n} & D_v & \mathbf{0}_{n \times n} & \mathbf{0}_{n \times 1} \\[4pt]
	\mathbf{0}_{n \times n} & \mathbf{0}_{n \times n} & D_\lambda & \mathbf{0}_{n \times 1} \\[4pt]
	\mathbf{0}_{1 \times n} & \mathbf{0}_{1 \times n} & \mathbf{0}_{1 \times n} & \nu_r\sqrt{r_t} \\[4pt]
D_Y& \mathbf{0}_{1 \times n} & \mathbf{0}_{1 \times n} & 0
\end{pmatrix}  ,
\end{equation*}
where $
D_v = 	\text{diag}\left( \nu_v^1\sqrt{v^1_t}, \dots, \nu_v^n\sqrt{v^n_t} \right)_{n\times n},$ $
D_\lambda = \text{diag}\left( \nu_\lambda^1\sqrt{\lambda^1_t}, \dots, \nu_\lambda^n\sqrt{\lambda^n_t} \right)_{n\times n}$, $
D_Y =	\left( \pi_t^1 Y^\alpha_{t-}\sqrt{v^1_t}, \dots, \pi_t^n Y^\alpha_{t-}\sqrt{v^n_t} \right)_{1\times n}$.
Therefore, the value function
\begin{equation*}
	V(t,x) = \sup_\alpha \mathbb{E}\left[\int_{t}^{T}e^{-\rho(s-t)}\frac{\left(c_{s}Y^\alpha_{s}\right)^{\delta}}{\delta}ds+e^{-\rho(T-t)}\frac{\left(Y^\alpha_{T}\right)^{\delta}}{\delta} \, \Big| \, X^\alpha_t =x \right] ,
\end{equation*}
satisfies the following HJB equation for all $(t,x) \in [0,T) \times \mathbb{R}^n_+ \times \mathbb{R}^n_+ \times \mathbb{R} \times \mathbb{R}_+$, 
\begin{align}
	0 = \hspace{0.75mm}& \partial_t V(t,x) - \rho V(t,x) + \sup_{(c,\pi) \in A} \bigg\{ \mu_X^\top\hspace{-0.2mm}(t,x) \, \nabla_x V(t,x)  +\frac{1}{2} \text{Tr}\left[\Sigma_X(t,x) \Sigma \Sigma_X^\top(t,x) \, \nabla^2_xV(t,x)  \right] \nonumber
	\\
	&\hspace{6.6mm}  + \sum_{i=1}^{n} \lambda^i \, \mathbb{E}\left[V(t, v, \lambda, r, y+y \, \pi^{i} (e^{Z^i_1} - 1)) - V(t,x) \right] +\frac{(cy)^\delta}{\delta}\bigg\} \, , \label{HJBinv2}
\end{align}
with terminal condition $V(T, x) = y^\delta/\delta$. We consider a portfolio of $n=25$ stocks, resulting in a $52$-dimensional value function and a $26$-dimensional control process, with the same parameters as in Section \ref{optcv}. This version of the consumption-investment problem is significantly more complex than the one discussed in Section \ref{optcv}, with the value function's dimensionality increasing from 2 to 52. Consequently, Runge--Kutta can no longer be used to solve the associated HJB equation \eqref{HJBinv2}, making it impossible to compute the ${\rm MAE}_V$ and ${\rm MAE}_\alpha$ metrics. However, despite this lack of a reference for comparison, GPI-CBU still produces results that appear reasonable. We indeed first observe in Figure \ref{fig:main4} (main manuscript) that the losses $\widehat{\mathscr{L}}^{(k)}_1$ \eqref{How3} and  $\widehat{\mathscr{L}}^{(k)}_2$ \eqref{loss4} converge as the number of epoch $k$ increases. Note that the orange curve in the left plot represents the interior loss (first right-hand term) of  $\widehat{\mathscr{L}}^{(k)}_1$, while the blue curve is the boundary loss (second right-hand term) of $\widehat{\mathscr{L}}^{(k)}_1$, see Eq.\ \eqref{How3}.

The following Figures \ref{comp14} and \ref{comp15} confirm that the results of GPI-CBU for both the value function and optimal control are reasonable, as they closely resemble those of Figures \ref{comp4} and \ref{comp5}. For the value functions of Figure \ref{comp14},  the standard deviations across 10 independent runs of GPI-CBU remain very low, confirming the stability of the approximations. For the optimal consumption rate in Figure \ref{comp15} (left plot), the standard deviation across the 10 runs tends to be higher for values of time $t$ close to 0, although being still reasonable. Finally,  varying the first dimension of the intensity $\lambda^1$ mainly impacts the corresponding fraction of wealth $\pi^{*,1}_t$, while its effect on the other proportions and consumption rate is more moderate (since arising from the correlation matrix $\Sigma$ of the Brownian motion $W$).
\begin{figure}[H]
	\centering
	\begin{minipage}{.5\textwidth}
		\includegraphics[width=0.97\linewidth]{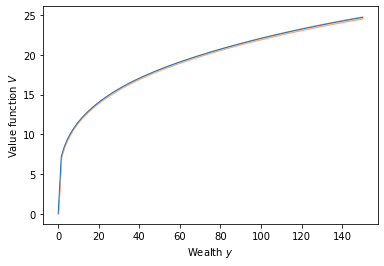}
		\label{fig:test14b}
	\end{minipage}%
	\begin{minipage}{.5\textwidth}
		\includegraphics[width=0.97\linewidth]{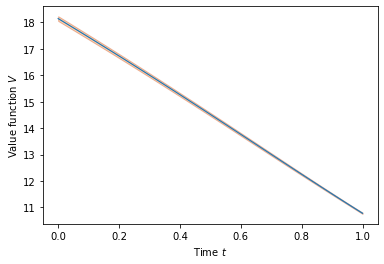}
		\label{fig:test15b}
	\end{minipage}
\caption{\small Value function $V(t,x)$ for $t=0$ and $x= (0.15 \cdot \mathbf{1}_{10}, \mathbf{1}_{10}, 0.02, y)$ with $y \in [0,150]$ (left) and for $t \in [0,1]$ and  $x= (0.15 \cdot \mathbf{1}_{10},  \mathbf{1}_{10}, 0.02, 50)$ (right), for the optimal consumption-investment problem \eqref{inv1} with stochastic coefficients and $n = 25$ stocks. Blue lines:  results of GPI-CBU with $\pm$ 1 standard deviation given by orange shaded area ($k_{*}= 1000$).} \vspace{-3mm}
	\label{comp14}
\end{figure} \vspace{-1mm}
\begin{figure}[H]
	\centering
	\begin{minipage}{.5\textwidth}
		\includegraphics[width=0.97\linewidth]{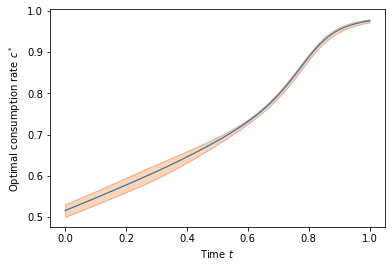}
		\label{fig:test14}
	\end{minipage}\hspace{-1mm}%
	\begin{minipage}{.5\textwidth} \vspace{0.8mm}
		\includegraphics[width=0.97\linewidth]{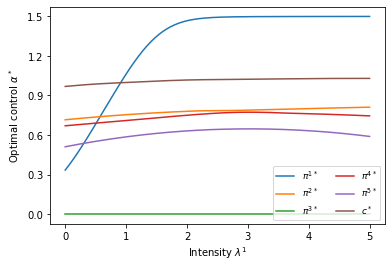}
		\label{fig:test15}
	\end{minipage}
\caption{\small Optimal policy for the optimal consumption-investment problem \eqref{inv1} with stochastic coefficients and $n = 25$ stocks. Left plot: optimal consumption rate $c^*(t,x)$ for $t\in [0,1]$ and $x= (0.15 \cdot \mathbf{1}_{10},  \mathbf{1}_{10}, 0.02, 50)$, results of  GPI-CBU in blue and $\pm$ 1 standard deviation given by orange shaded area ($k_{*}= 1000$). Right plot: optimal consumption rate $c^*(t,x)$ and first five optimal fractions of wealth $\pi^*(t,x)$ for  $t=0$ and $x= (0.15 \cdot \mathbf{1}_{10}, \lambda^1, \mathbf{1}_{9}, 0.02, 50)$ for $\lambda^1\in [0,5]$.}
	\label{comp15}
\end{figure}

\newpage

\end{document}